\PassOptionsToPackage{prologue,table,xcdraw}{xcolor}
\documentclass[sigconf,screen]{acmart}
\AtBeginDocument{%
  \providecommand\BibTeX{{%
    \normalfont B\kern-0.5em{\scshape i\kern-0.25em b}\kern-0.8em\TeX}}}

\copyrightyear{2023}
\acmYear{2023}
\setcopyright{acmlicensed}
\acmConference[MM '23] {Proceedings of the 31st ACM International Conference on Multimedia}{October 29--November 3, 2023}{Ottawa, ON, Canada.}
\acmBooktitle{Proceedings of the 31st ACM International Conference on Multimedia (MM '23), October 29--November 3, 2023, Ottawa, ON, Canada}
\acmPrice{15.00}
\acmISBN{979-8-4007-0108-5/23/10}
\acmDOI{10.1145/3581783.3611932}

\usepackage{subcaption}

\usepackage{amssymb}
\usepackage{enumitem}
\usepackage{multirow}
\usepackage{bbding}
\usepackage{soul}
\usepackage[table,xcdraw]{xcolor}
\usepackage{balance}
\begin{document}

\title{Fine-Grained Spatiotemporal Motion Alignment for Contrastive Video Representation Learning}

% \affiliation{%

% \affiliation{%

% \affiliation{%

% \author{Minghao Zhu}
% \affiliation{%
%  \institution{Tongji University}
%  \streetaddress{4800 Caoan Highway}
%  \city{Jiading Qu}
%  \state{Shanghai}
% \email{tjdezmh@tongji.edu.cn}

\author{Minghao Zhu}
\email{zmhh_h@tongji.edu.cn}
\affiliation{%
 \institution{Tongji University}
 \streetaddress{4800 Caoan Highway}
 \city{}
 \state{}
 \country{}}

\author{Xiao Lin}
\email{linx_xx@tongji.edu.cn}
\affiliation{%
 \institution{Tongji University}
 \streetaddress{4800 Caoan Highway}
 \city{}
 \state{}
 \country{}}

\author{Ronghao Dang}
\email{dangronghao@tongji.edu.cn}
\affiliation{%
 \institution{Tongji University}
 \streetaddress{4800 Caoan Highway}
 \city{}
 \state{}
 \country{}}

\author{Chengju Liu}
\authornote{Corresponding author.}
\email{liuchengju@tongji.edu.cn}
\affiliation{%
 \institution{Tongji University}
 \streetaddress{4800 Caoan Highway}
 \city{}
 \state{}
 \country{}}

\author{Qijun Chen}
\email{qjchen@tongji.edu.cn}
\affiliation{%
 \institution{Tongji University}
 \streetaddress{4800 Caoan Highway}
 \city{}
 \state{}
 \country{}}

\renewcommand{\shortauthors}{Minghao Zhu, Xiao Lin, Ronghao Dang, Chengju Liu, \& Qijun Chen}

%%%%%%%%% ABSTRACT
\begin{abstract}
  As the most essential property in a video, motion information is critical to a robust and generalized video representation. To inject motion dynamics, recent works have adopted frame difference as the source of motion information in video contrastive learning, considering the trade-off between quality and cost. However, existing works align motion features at the instance level, which suffers from spatial and temporal weak alignment across modalities. In this paper, we present a \textbf{Fi}ne-grained \textbf{M}otion \textbf{A}lignment (FIMA) framework, capable of introducing well-aligned and significant motion information. Specifically, we first develop a dense contrastive learning framework in the spatiotemporal domain to generate pixel-level motion supervision. Then, we design a motion decoder and a foreground sampling strategy to eliminate the weak alignments in terms of time and space. Moreover, a frame-level motion contrastive loss is presented to improve the temporal diversity of the motion features. Extensive experiments demonstrate that the representations learned by FIMA possess great motion-awareness capabilities and achieve state-of-the-art or competitive results on downstream tasks across UCF101, HMDB51, and Diving48 datasets. Code is available at \url{https://github.com/ZMHH-H/FIMA}.
\end{abstract}




% \received{23 April 2023}

\maketitle

%%%%%%%%% BODY TEXT
%------------------------------------------------------------------------
\begin{figure}[t]
\begin{center}
\includegraphics[width=\linewidth]{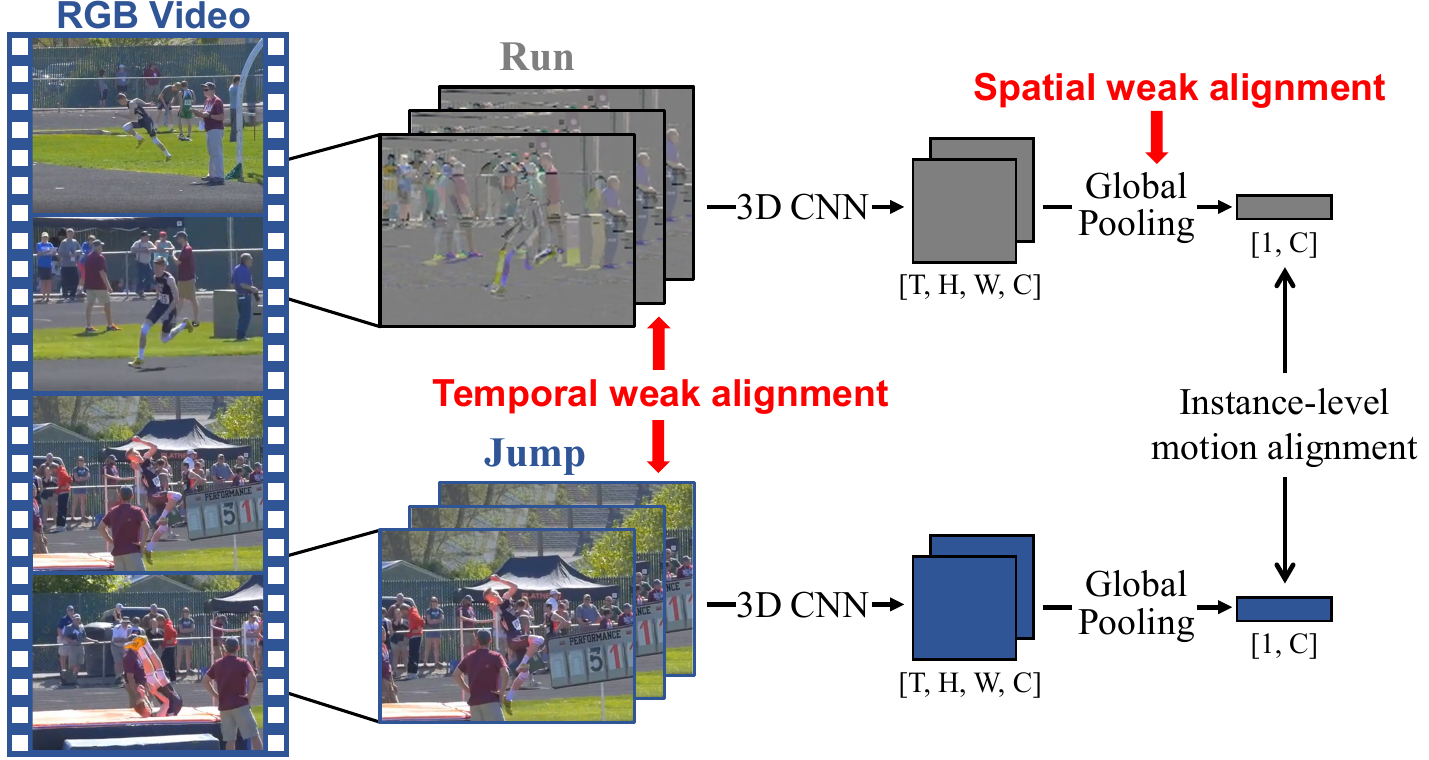}
\end{center}
% \setlength{\belowcaptionskip}{-0.3cm}
% \vspace{-0.4cm}
   % \caption{An illustration of weak alignment between RGB and frame difference. For temporal weak alignment, since the motion semantics varies over time, direct alignment of motion features with different timestamps introduces minimal shared information between views. For spatial weak alignment, the pooled feature is distracted by cluttered background noise, leading to the misalignment of background features and inadequate alignment of foreground features.}
   \caption{An illustration of weak alignment across modalities. For temporal weak alignment, since motion semantics varies over time, direct alignment of features with different timestamps introduces minimal shared information. For spatial weak alignment, the pooled feature is distracted by cluttered background noise, leading to misalignment of background features and inadequate alignment of foreground features.}
\label{Motivation}
\end{figure}

\section{Introduction}
\label{section:sec1}

With the enormous growth of uncurated data on the Internet, self-supervised learning came with the promise of learning powerful representations from unlabelled data that can be transferred to various downstream tasks. In particular, contrastive self-supervised learning based on instance discrimination~\cite{InstDisc} has achieved great success in both NLP~\cite{Bert,GPT3} and computer vision~\cite{SimCLR,MoCov1,CLIP}. In the video domain, this learning diagram has also presented promising performance by keeping the instances within the same video semantically consistent~\cite{ALargeScale,CVRL}. However, vanilla video contrastive learning has difficulty modeling local temporal information~\cite{TCLR,Static} and possesses severe background bias~\cite{FAME,BE}, which limits the generalization and transferability of the learned representations. The reason may stem from the existence of static bias in the positive pair construction.
% The model could distinguish temporally separated instances by attending to the static cues while neglecting the dynamic details which provide crucial information for discrimination and downstream tasks.
% The model could pull the features of temporally separated instances close by attending to the static cues while neglecting the dynamic details which provide crucial information for discrimination and downstream tasks.
The features of temporally separated instances could be easily pulled close by attending to the static cues while neglecting the dynamic details, which provide crucial information for discrimination and downstream tasks.

To alleviate the background bias in the context of contrastive learning, an effective method is to introduce motion information. Previous works incorporate motion information either by constructing motion-prominent positive pairs via meticulously designed data augmentations~\cite{FAME,DSM,BE} or using other modalities to explicitly enhance motion information in feature space~\cite{Mfocused,Msensitive,MACLR}. Among them, aligning the motion features of optical flow in an explicit way achieves impressive results. But the expensive computation cost limits the scalability of optical flow on large-scale datasets. Hence, how to incorporate motion information effectively without resorting to costly accessories has attracted a lot of attention.

Frame difference, an alternative with a negligible cost for optical flow, can extract motion across frames by removing the still background. But its quality is extremely susceptible to background noise caused by camera jitter or drastic changes in the background. Several approaches~\cite{Dual,Static,interintra,improvedIIC} employ it in video contrastive learning and show promising results. Even with notable improvements in performance, how to effectively align the RGB and frame difference features in the latent space still not be fully explored. 

The previous works align the features of RGB and frame difference from temporally different views by using global feature vectors~\cite{Dual,improvedIIC}, which may suffer from the $\mathit{weak~alignment}$ between the two modalities. 
As illustrated in Fig.~\ref{Motivation}, we summarize and categorize the weak alignment problem into two types: $\mathtt{temporal~weak~alignment}$ and $\mathtt{spatial~weak~alignment}$.
Temporal weak alignment signifies that two clips at different timestamps are not semantically consistent in the concept of motion. Some action classes consist of multiple stages of sub-motions. For example, high jumping in Fig.~\ref{Motivation} has two motion stages: running and jumping. These two stages have very different motion semantics, and directly pulling them close in the feature space will make the model invariant to the motion semantic changes along the temporal dimension. 
Spatial weak alignment can be viewed as an inherent problem of the instance-level contrastive learning framework. The global average pooling extends the receptive field of the features to the entire view and compresses all information into a one-dimensional vector, resulting in the pooled features losing spatial information and being distracted by background noise. Simply aligning the pooled features may lead to the misalignment of background features and inadequate alignment of foreground features. As shown in Fig.~\ref{Motivation}, since the frame difference contains a lot of background noise, the alignment of pooled features becomes sub-optimal for introducing significant motion information.
% Spatial weak alignment means inconsistent focus over space. It happens when one of the features is distracted by background noise. Further, the global average pooling extends the receptive field of the features to the entire view, leading to the misalignment of background features and inadequate alignment of foreground features. As shown in Fig.~\ref{Motivation}, since the feature map of frame difference incorrectly attends to the background area, the pooled feature becomes background dominant. Consequently, the background noise are introduced while the useful motion features are neglected.

In this paper, we present a novel framework to introduce well-aligned motion features in video contrastive learning, namely \textbf{Fi}ne-grained \textbf{M}otion \textbf{A}lignment (FIMA). 
We argue that the alignment of motion features should be at a more fine-grained level. To this end, we discard the global average pooling and construct a pixel-level contrastive learning framework, where each pixel of the RGB feature map tries to predict the motion feature pixels at the same spatial location but the different timestamps. In this way, the alignments of foreground and background features are sufficient and decoupled. Based on it, we address the temporal weak alignment by designing a motion decoder that takes the RGB feature map from the target view as the bridge between temporally distant positive pairs. This task requires the model to not only learn the correct correspondence between RGB features from the target and the source view but also encode enough motion information for cross-modality reconstruction. Thus, spatiotemporally discriminative motion features can be learned. To tackle the spatial weak alignment, we propose a foreground sampling strategy to filter out the background pixels in the construction of positive pairs, hence avoiding the distraction of background noise. In addition, we further propose a frame-level motion reconstruction task for improving the temporal diversity of the motion features. Given a frame of RGB feature map, the motion decoder learns to reconstruct the exactly overlapped local motion feature and distinguish it from others.
We summarize our main contributions as follows:
\begin{itemize}[noitemsep,topsep=0pt,leftmargin=*]
% \begin{itemize}[noitemsep,topsep=0pt]
\item We demonstrate the weak alignment problem between RGB and frame difference modalities, and analyze the influence of the weak alignment in terms of time and space.
\item We present a novel FIMA framework, consisting of a dense contrastive learning paradigm, a foreground sampling strategy, and a motion decoder to eliminate the weak alignment of two modalities at the pixel and the frame level, which enhances the self-supervised representations.
\item Our framework achieves state-of-the-art or competitive results on two downstream tasks, action recognition, and video retrieval, across UCF-101, HMDB-51, and Diving-48 datasets.
\end{itemize}
%------------------------------------------------------------------------
\section{Related Work}
\noindent
{\bf Self-supervised learning in videos.} Self-supervised learning aims to learn transferable representations from large-scale unlabeled data. In the video domain, early works focus on encoding intrinsic structure information by solving sophisticated designed pretext tasks, including temporal transformation~\cite{speednet,shufflearn,arrowtime,cliporder,PRP,TempTrans}, statistics prediction~\cite{geometryguided,appearancestatistics}, and spatiotemporal puzzles~\cite{cubicpuzzles,videocloze}. Later, contrastive learning based on instance discrimination~\cite{InstDisc} makes great progress in the image domain~\cite{SimCLR,BYOL,MoCov1}. Some works extend it into video representation learning and achieve promising results~\cite{ALargeScale,CVRL}. To further enhance video representations, a line of works construct various positive views by applying spatiotemporal augmentations~\cite{LongShort,GlobalContext,FAME,VideoMoCo,MultiLevel}. Besides, ~\cite{RSPNet,ASCNet,contrastivepretext} formulate predictive tasks in a contrastive manner. ~\cite{AVclustering,SLIC} leverage the idea of clustering in video contrastive learning. ~\cite{AVclustering,VideoText,Broaden} seek consistency between multi-model data, such as audio, text, and optical flow.

\noindent
{\bf Alleviation of background bias in videos.} Background bias in commonly used datasets may lead to the model over-focusing on static cues, resulting in poor generalization. To overcome this, ~\cite{Diving48} proposes a procedure to assemble a less biased dataset. In the context of video self-supervised learning, DSM~\cite{DSM}, BE~\cite{BE}, and FAME~\cite{FAME} implicitly mitigate the background bias by constructing motion-prominent positive or negative samples. ~\cite{MACLR,Mfocused,Msensitive} explicitly introduce motion information into feature space from other modalities like optical flow. DCLR~\cite{Dual} uses frame difference as motion supervision and decouples it from data input and feature space. In this paper, we address the background bias by selectively introducing significant motion information from frame difference. 

\noindent
{\bf Dense supervision in contrastive learning.} Dense contrastive learning is initially devised for dense prediction tasks in the image domain~\cite{DenseCL,PixPro}, such as object detection and semantic segmentation. It preserves spatial information by constructing pixel-level positive pairs between dense features from different views. 
% and closes the gap between self-supervised learning and dense prediction tasks. 
In the video domain, some works exploit dense supervision in contrastive learning. ~\cite{DensePredictiveCoding,MemoryAug,ContextAndMotionDecoup} propose to predict dense feature maps in the future. They generate positive samples by applying consistent geometric transformation across a video. However, such a strategy can not induce sufficient occlusion invariance, which has been proven to be crucial for contrastive representation learning~\cite{Demystifying}. ~\cite{Contextualized} extends~\cite{DenseCL} to the video domain and constructs region-level positive pairs by calculating the correspondence between local RGB features. But this strategy hardly establishes a correct correspondence between RGB and frame difference due to the discrepancy between the two modalities. To address these limitations, our method extends~\cite{PixPro} to the spatiotemporal domain and constructs positive pairs based on spatial location prior.

%------------------------------------------------------------------------
%------------------------------------------------------------------------
\begin{figure*}[htb]
\begin{center}
\includegraphics[width=\textwidth]{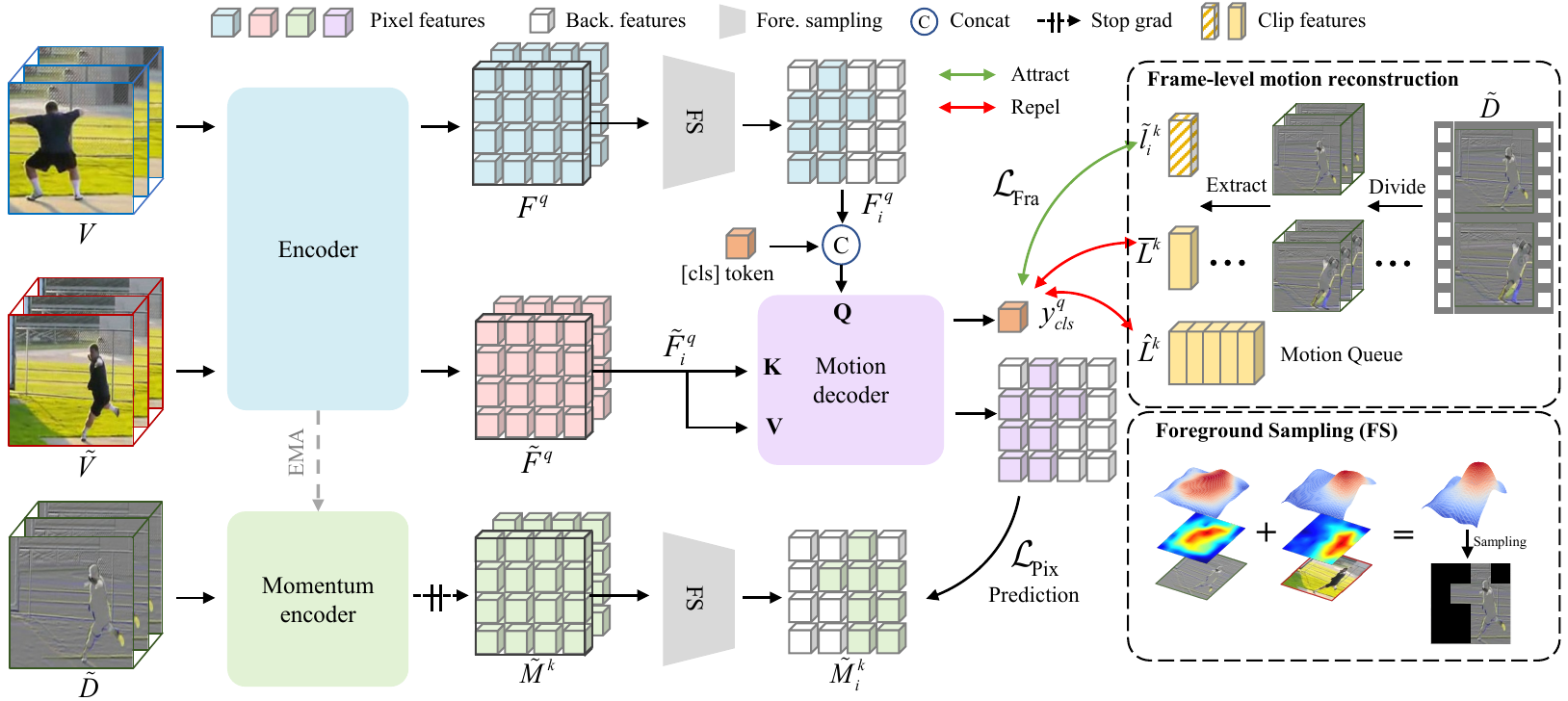}
\end{center}
% \vspace{-0.4cm}
\caption{Overview of the framework. We sample two temporally distant clips $\{V, \Tilde{V}\}$ and compute the frame difference $\Tilde{D}$. The corresponding dense feature maps $\{F^q, \Tilde{F}^q, \Tilde{M}^k\}$ are extracted by the encoder or its momentum version. We sample the foreground features at the $i$-th frame of $F^q$ and concatenate them with a class token, then feed them into the motion decoder. We use the motion decoder to reconstruct the foreground features of $\Tilde{M}^k$ in the $i$-th frame, by collecting information from $\Tilde{F}_i^q$. Finally, the class token is used to reconstruct the local motion feature with time interval overlaps exactly with $\Tilde{F}_i^q$. 
}
\label{framework}
\end{figure*}
%------------------------------------------------------------------------
\section{Method}

 An overview of our proposed framework is presented in Fig.~\ref{framework}. In Section~\ref{section:sec3.1}, we first revisit vanilla spatiotemporal contrastive learning and extend it to the pixel level to generate dense motion supervision. In Section~\ref{section:sec3.2}, we elaborate on the motion decoder and foreground sampling strategy designed to eliminate temporal and spatial weak alignment. Finally, we present the frame-level motion reconstruction task to improve the feature diversity in Section~\ref{section:sec3.3}. 

\subsection{Pixel-level Contastive Learning in Videos} 
\label{section:sec3.1}

Given a video, the vanilla instance-level spatiotemporal contrastive learning randomly samples two video clips $\big\{V, \Tilde{V}\big\}$ at different timestamps. The clips are augmented by temporally consistent augmentation~\cite{CVRL} and processed by a feature encoder with the global average pooling to extract corresponding video-level representations $\big\{\nu, \Tilde{\nu}\big\}$. The prevalent InfoNCE loss~\cite{CPC} is adopted for optimization:
\begin{equation} \label{eq1}
\mathcal{L}_{\mathrm{VV}}=-\log \frac{h\big(\nu^q, \Tilde{\nu}^k\big)}{h\big(\nu^q, \Tilde{\nu}^k\big)+\sum_{\hat{\nu}^k} h\big(\nu^q, \hat{\nu}^k\big)},
\end{equation}
% $$\mathcal{L}_{\mathrm{VD}}=-\log \frac{h\left(\nu^q, \nu^k\right)}{h\left(\nu^q, \nu^k\right)+\sum_{k} h\left(\nu^q, \hat{\nu}^k\right)},$$
% $$\mathcal{L}_{\mathrm{VD}}=-\log \frac{\exp\left(\operatorname{sim}\left(\nu^q, \nu^k\right)/\tau\right)}{h\left(\nu^q, \nu^k\right)+\sum_{k} h\left(\nu^q, \bar{\nu}^k\right)},$$
% $$\mathcal{L}_{\mathrm{VD}}=-\log \frac{h\left(\nu_{\mathrm{glb}}^q, \nu_{\mathrm{glb}}^k\right)}{h\left(\nu_{\mathrm{glb}}^q, \nu_{\mathrm{glb}}^k\right)+\sum_{i=1}^N h\left(\nu_{\mathrm{glb}}^q, \bar{\nu}_{i, \mathrm{glb}}^k\right)},$$
where $h(x, y)=\exp \big(g(x)^T g(y) /\|g(x)\|\|g(y)\| \tau\big)$ measures the similarity between two projected feature vectors $g(x)$ and $g(y)$; $g(\cdot)$ is a non-linear projection head network; $\tau$ is the temperature parameter; Negative keys $\hat{\nu}^k$ are taken from a memory bank as we follow the network design of MoCo~\cite{MoCov1}. Note that the superscripts $q$ and $k$ indicate the features extracted by the query encoder and the momentum encoder, respectively. This training objective aims to pull the query and its positive key closer while it repels other negative keys in the latent space. However, the representations learned based on Eq.~\eqref{eq1} are easily overwhelmed by static background cues~\cite{FAME,BE} and lack the ability to capture dynamic motion information~\cite{TCLR}. To introduce motion information, previous works~\cite{Dual,improvedIIC} use frame difference as the other motion representation learning branch. We calculate the frame difference $\Tilde{D}$ by differentiating adjacent frames of $\Tilde{V}$ and extract the motion feature $\Tilde{d}^k$. The motion contrastive loss is as follows: 
\begin{equation}
\mathcal{L}_{\mathrm{VD}}=-\log \frac{h\big(\nu^q, \Tilde{d}^k\big)}{h\big(\nu^q, \Tilde{d}^k\big)+\sum_{\hat{d}^k} h\big(\nu^q, \hat{d}^k\big)},
\end{equation}
% where $\hat{d}^k$ denotes the motion features of other videos.
where $\hat{d}^k$ is the motion features of other videos in a memory bank. 

As mentioned in Section~\ref{section:sec1}, the globally average-pooled features cannot be well-aligned due to the weak alignment between modalities. Thus, we extend the dense contrastive learning framework~\cite{PixPro} to the spatiotemporal domain to generate dense motion supervision. The representations extracted from $\big\{V, \Tilde{D}\big\}$ are kept as feature maps, denoted as $\big\{F^q, \Tilde{M}^k\big\} \in \mathbb{R}^{T \times H \times W \times C}$, where $T,H,W,C$ are the dimensions of the time, height, width, channel, respectively. $F_i^q$ and $\Tilde{M}_i^k$ represent the $i$-th frame of the feature maps. For each feature pixel $f_i^q \in F_i^q$, we assign a positive pixel set $\phi_p^k \subseteq \Tilde{M}_i^k$ based on the spatial location prior. Specifically, we record the original coordinates of the clip when applying the geometric transformation (e.g. crop and flip). Then, each feature pixel in $F_i^q$ and $\Tilde{M}_i^k$ is warped to the original video spatial space, and the two-dimensional (height and width) Euclidean distances between $f_i^q$ and all pixels in $\Tilde{M}_i^k$ are computed. The distances are normalized to the diagonal length of a feature map bin and a hyperparameter $\mathcal{T}$ is used to measure the distance in scale. We set the threshold $\mathcal{T}=0.7$ by default. The pixels in $\Tilde{M}_i^k$ with a distance smaller than $\mathcal{T}$ are assigned to the $\phi_p^k$. The rest of the pixels in the feature map $\Tilde{M}^k$ and motion feature pixels from other videos are assigned to negative pixel set $\phi_n^k$. The pixel-level motion contrastive loss can be formulated as:
\begin{equation} \label{eq3}
\mathcal{L}_{\mathrm{Pix}}=-\log \frac{\sum\limits_{\Tilde{m}_i^k \in \phi_p^k} h\big(f_i^q, \Tilde{m}_i^k\big)}{\sum\limits_{\Tilde{m}_i^k \in \phi_p^k} h\big(f_i^q, \Tilde{m}_i^k\big)+\sum\limits_{\hat{m}_i^k \in \phi_n^k} h\big(f_i^q, \hat{m}_i^k\big)}.
\end{equation}
Note that the projection head $g(\cdot)$ here is instantiated as two successive $1 \times 1$ convolution layers to adapt the input form of the feature map. The loss is averaged over all feature pixels in $F^q$ with at least one positive pair (i.e., corresponding $\phi_p^k \ne \varnothing$). Intuitively, given an RGB feature pixel in the source view, the network learns to predict the motion features at the same spatial location in the target view.

\subsection{Fine-grained Motion Feature Alignment}\label{section:sec3.2}

The pixel-level $\mathcal{L}_{\mathrm{Pix}}$ constructs a framework for fully utilizing dense motion supervision. Based on it, we propose to separately eliminate the weak alignment of motion features in terms of time and space.\\
{\bf Temporal Weak Alignment Elimination.} The positive pair constructed by $\mathcal{L}_{\mathrm{Pix}}$ still exists two limitations: the first is that the loss pulls features at different timestamps close, resulting in the model becoming invariant to the motion semantic change along the temporal dimension; the second is that the receptive field of a feature pixel only covers a limited region, which may lead to a greater semantic discrepancy between positive pairs.

To address these limitations, we propose to use the RGB features $\Tilde{F}^q$ extracted from the target view $\Tilde{V}$ by the query encoder as the bridge between $F^q$ and $\Tilde{M}^k$. To this end, we design a motion decoder based on the attention mechanism, which is quite effective in spatiotemporal dependence modeling~\cite{Non-local,Contextualized,ViViT,TimeSformer}.
Specifically, we consider the dense features $\Tilde{F}_i^q$ as a collection containing various information. For each RGB feature pixel $f_i^q \in F_i^q$, the motion decoder tries to reconstruct the motion features at the same spatial location in the target view by querying the information in collection $\Tilde{F}_i^q$. Then the pixel-level motion contrastive loss becomes:
\begin{align}
& y_i^q = \mathrm{MD}(f_i^q,\Tilde{F}_i^q,\Tilde{F}_i^q), \\
& \mathcal{L}_{\mathrm{Pix}} = -\log \frac{\sum\limits_{\Tilde{m}_i^k \in \phi_p^k} h\big(y_i^q, \Tilde{m}_i^k\big)}{\sum\limits_{\Tilde{m}_i^k \in \phi_p^k} h\big(y_i^q, \Tilde{m}_i^k\big)+\sum\limits_{\hat{m}_i^k \in \phi_n^k} h\big(y_i^q, \hat{m}_i^k\big)},
\end{align}
where $\mathrm{MD}(Q,K,V)$ refers to the motion decoder implemented by the standard transformer layer~\cite{attention}; $Q \in \mathbb{R}^{1 \times d}$ is the query feature; $K,V \in \mathbb{R}^{HW \times d}$ are key-value pairs; $d$ is the query/key dimension; $y_i^q$ is the output of the motion decoder.
% This task requires the extracted feature maps to be sufficiently good, such that the correct correspondence between feature pairs in $F_i^q$ and $\Tilde{F}_i^q$ can be built to query meaningful information. 
The motion decoder builds the correct correspondence between the RGB features of the source view $F_i^q$ and the target view $\Tilde{F}_i^q$ and avoids enforcing the features at different timestamps to be similar.
It also requires every RGB feature pixel in $\Tilde{F}_i^q$ to encode more motion information around itself for cross-modality reconstruction. Further, the attention operation extends the receptive field of the predicted feature pixels to the entire target view, thus eliminating the semantic discrepancy. \\
{\bf Spatial Weak Alignment Elimination.} 
With the proposed dense contrastive learning framework, the alignment of each foreground and background feature pixel is decoupled. We can avoid the disturbance of noise by filtering out the background pixels.
% Since the proposed pixel-level contrastive learning framework fully aligns both foreground and background features, we can avoid the disturbance of noise by filtering out the background pixels. 
As a common visualization technique, class activation maps~\cite{CAAM,CAM} provide an intuitive way to localize the discriminative regions, by which we expect to classify the foreground and background pixels.
However, we find that activation maps from either RGB or frame-difference features easily attend to the incorrect background area. 
The other observation is that the distribution of the activation map is relatively flat when the model's attention is disturbed by background noise. In other words, it tends to cover a larger range of the background with lower activation values. Instead, the distribution of the activation map is steeper when the model correctly captures the foreground information. The underlying reason lies in the foreground information being more structural and concentrated, producing a distinct and dense activation region. 
Based on this observation, we propose to use the features of two modalities with complementary information to jointly determine the foreground region.
% This observation suggests that aggregating the features of two modalities with complementary information to jointly determine the foreground region is more accurate than using a single modality.
% Concretely, we divide the pixels in the target view $\Tilde{M}^k$ into two mutually exclusive sets $\Tilde{M}_{fg}^k$ and $\Tilde{M}_{bg}^k$, where $fg$ and $bg$ mean foreground and background. We compute class-agnostic activation maps~\cite{CAAM} of two modalities by applying the average pooling along channel and time dimensions, then fuse them by simple point-wise addition. We assign pixels to the $\Tilde{M}_{fg}^k$ and the $\Tilde{M}_{bg}^k$ based on the fused activation map $\Tilde{Z}^k \in \mathbb{R}^{H \times W}$ as follows: 
Concretely, we compute class-agnostic activation maps~\cite{CAAM} of two modalities by applying the average pooling along channel and time dimensions, then fuse them by simple point-wise addition. We divide pixels in the target view $\Tilde{M}^k$ into two mutually exclusive sets $\Tilde{M}_{fg}^k$ and $\Tilde{M}_{bg}^k$ based on the fused activation map $\Tilde{Z}^k \in \mathbb{R}^{H \times W}$ as follows: 
\begin{equation} \label{eq6}
A\big(\Tilde{m}_{i,j}^k\big)=
\begin{cases}1, & \text { if } \Tilde{z}_j > \beta \text{-th quantile of } \Tilde{Z}^k, \\ 0, & \text{otherwise}, \end{cases}
\end{equation}
where $\Tilde{m}_{i,j}^k$ is a pixel in the feature map $\Tilde{M}^k$; $i$ denotes the temporal index and $j \in \{(1,1),(1,2),...,(H,W)\}$ is the spatial index; $\Tilde{z}_j$ is the pixel at spatial index $j$ in the fused activation map; $\beta \in [0,1]$ is a hyper-parameter used to describe the portion of the foreground. We set $\beta=0.5$ by default. Similarly, we obtain the foreground set $F_{fg}^q$ and the background set $F_{bg}^q$
 of the source view $F^q$ in the same manner. We only sample positive pairs in the foreground regions and the pixel-level motion contrastive loss becomes:
\begin{align}
& y_i^q = \mathrm{MD}(f_i^q,\Tilde{F}_i^q,\Tilde{F}_i^q), \\
& \mathcal{L}_{\mathrm{Pix}} = -\log \frac{\sum\limits_{\Tilde{m}_i^k \in \phi_p^k \cap \Tilde{M}_{fg}^k} h\big(y_i^q, \Tilde{m}_i^k\big)}{\sum\limits_{\Tilde{m}_i^k \in \phi_p^k \cap \Tilde{M}_{fg}^k} h\big(y_i^q, \Tilde{m}_i^k\big)+\sum\limits_{\hat{m}_i^k \in \phi_n^k} h\big(y_i^q, \hat{m}_i^k\big)},
\end{align}
where $\phi_n^k$ here indicates the rest of the pixels in the feature map $\Tilde{M}^k$ except $\big\{\phi_p^k \cap \Tilde{M}_{fg}^k\big\}$, plus motion feature pixels from other videos. The loss is averaged over all feature pixels in the foreground set $F_{fg}^q$ with at least one positive pair.

\subsection{Frame-level Motion Feature Reconstruction}\label{section:sec3.3}

The $\mathcal{L}_{\mathrm{Pix}}$ aligns the motion features at the pixel level. To further enhance the temporal diversity of the learned features, we propose a frame-level local motion reconstruction task. We divide the motion clip $\Tilde{D}$ into $T$ sub-clips, where $T$ is the time dimension of the corresponding feature map $\Tilde{M}^k$. Given a frame of the feature map $\Tilde{F}_i^q$, the motion decoder aims to reconstruct the feature of the sub-clip with time interval overlaps exactly with $\Tilde{F}_i^q$. 

Before input to the motion decoder, we prepend a learnable $\texttt{[cls]}$ token to the sequence of features. For the output state of the class token $y_{cls}^q$ and the corresponding local motion feature $\Tilde{l}_i^k$, the frame-level motion contrastive loss can be formulated as:
\begin{align}\label{eq9}
& y_{cls}^q = \mathrm{MD}(\texttt{[cls]},\Tilde{F}_i^q,\Tilde{F}_i^q), \\\label{eq10}
& \mathcal{L}_{\mathrm{Fra}}=-\log \frac{h\big(y_{cls}^q, \Tilde{l}_i^k\big)}
{h\big(y_{cls}^q, \Tilde{l}_i^k\big)+\sum\limits_{l^k \in \{\bar{L}^k,\hat{L}^k\}} h\big(y_{cls}^q, l^k\big)},
\end{align}
where $\bar{L}^k$ and $\hat{L}^k$ are the sets of local features in the same video and other videos. We use the motion decoder shared with $\mathcal{L}_{\mathrm{Pix}}$ to avoid introducing extra overhead.
By discriminating the positive from the intra-video negatives and inter-video negatives, the features extracted by the encoder become more temporally discriminative.
% To distinguish the corresponding local motion feature among others, the model needs to capture the subtle semantic variations by concentrating on semantically meaningful foreground rather than chaotic background noise.

The overall learning objective can be written as:
\begin{equation} \label{eq11}
\mathcal{L}=\mathcal{L}_{\mathrm{VV}}+\mathcal{L}_{\mathrm{Pix}} + \mathcal{L}_{\mathrm{Fra}},
\end{equation}
where we jointly optimize all losses and treat each term equally.
%------------------------------------------------------------------------
%----------TABLE ABLATION STUDY ON LOSS DESIGN-----------
\begin{table}[tbp]
\LARGE
\renewcommand\arraystretch{1.05}
\centering
\caption{Ablation study on the loss designs.}
\label{AblationLoss}
\resizebox{\linewidth}{!}{%
\begin{tabular}{cccccccc}
\hline
\multicolumn{4}{c}{Contrastive Losses} & \multicolumn{2}{c}{Linear} & \multicolumn{2}{c}{Finetune} \\ [-0.15cm]
$\mathcal{L}_{\mathrm{VV}}$        & $\mathcal{L}_{\mathrm{Pix}}$      & $\mathcal{L}_{\mathrm{Fra}}$      & $\mathcal{L}_{\mathrm{VD}}$       & UCF101       & HMDB51      & UCF101        & HMDB51       \\ \hline 
\Checkmark &           &           &           & 58.1         & 26.7          & 76.7          & 48.8         \\
\Checkmark &           &           & \Checkmark& 68.7         & 37.7          & 80.8          & 53.3         \\
\Checkmark & \Checkmark&           &           & 73.1         & 39.2          & 83.1          & 54.2         \\
\Checkmark &           & \Checkmark&           & 69.7         & 38.5          & 81.3          & 54.1         \\
\Checkmark & \Checkmark& \Checkmark&           &\textbf{75.3} & \textbf{42.8} & \textbf{84.2} & \textbf{57.8}         \\ \hline

\end{tabular}%
}
\end{table}
%---------------------------------------------------

\section{Experiments}

\subsection{Implementation Details}
\noindent
{\bf Datasets.} We conduct experiments on four standard video datasets, UCF101~\cite{UCF101}, HMDB51~\cite{HMDB51}, Kinetics-400~\cite{Kinetics}, and Diving-48~\cite{Diving48}. We use the updated V2 version of Diving-48 for evaluation.

\noindent
{\bf Technical Details.} We choose two widely used backbones, R(2+1)D-18~\cite{R(2+1)D}, and I3D-22~\cite{I3D} as the video encoder. The non-linear projection head is instantiated as two successive $1 \times 1$ convolution layers with an output dimension of 128 to adapt the input form of the feature map. We closely follow the network design of MoCo-v2~\cite{MoCov2}. Besides, the negative set $\phi_n^k$ in $\mathcal{L}_{\mathrm{Pix}}$ is implemented as a queue with a size of 784 for UCF101 and 31360 for Kinetics-400. We show more details in the supplementary material.

We implement the motion decoder by using the standard transformer layer in~\cite{attention}. As the default setting we use a 2-layer 512-wide model with 4 attention heads. The class token is a learnable 512-dim embedding. A linear layer is attached on two sides to adjust the feature dimension. The motion decoder is placed before the non-linear projection head. We add 1D absolute positional encodings to each feature frame before inputting it into the motion decoder.
% All tasks share the same weights of the video encoder and projection head.

\noindent
{\bf Self-supervised Pre-training.}
In the pre-training phase, we randomly sample two 16-frame clips with a temporal stride of 2 at different timestamps and compute the frame difference. Each clip is randomly cropped and resized to the size of $224 \times 224$ or $112 \times 112$ and then undergoes random horizontal flip, color jittering, random grayscale, and Gaussian blur in a temporally consistent way~\cite{CVRL}. We pretrain the model for 200 epochs on UCF101 or 100 epochs on Kinetics-400. Following the linear scaling rule of the learning rate~\cite{1hour}, we set the initial learning rate to 0.00375 with a total batch size of 24 for UCF101 or 0.01 with a total batch size of 64 for Kinetics-400. A half-period cosine schedule is used for the learning rate decay. We adopt SGD as the optimizer with a momentum of 0.9 and a weight decay of $10^{-4}$. 

\noindent
{\bf Downstream Task Evaluations.} We evaluate the self-supervised representations on two downstream tasks: action recognition and video retrieval. Following the common practice~\cite{MemoryAug,TCLR}, we average the predictions of 10 uniformly sampled clips of a video as the final result.
For action recognition, we use the weights of the pre-trained network as initialization and evaluate the representations under linear probing and fine-tuning settings. We report the top-1 classification accuracy on split 1 of UCF101 and HMDB51, and the v2 test set of Diving-48.
For video retrieval, we directly use the pre-trained model as a feature extractor without further training. Following~\cite{cliporder}, we extract the feature of each video in the test set as a query and retrieve the k-nearest neighbors in the training set by calculating the cosine similarity. We report the top-k recall R@k on UCF101 and HMDB51.

%----------TABLE ABLATION STUDY ON COMPONENTS-----------
\begin{table}[tbp]
\normalsize
\renewcommand\arraystretch{1.05}
\setlength{\tabcolsep}{6.3mm}
\centering

\caption{Ablation study on the components in $\mathcal{L}_{\mathrm{Pix}}$. Including dense contrastive learning framework (DC), foreground sampling strategy (FS), and motion decoder (MD). The first line indicates $\mathcal{L}_{\mathrm{Pix}}$ degrades to $\mathcal{L}_{\mathrm{VD}}$.}
\label{AblationComponents}
\resizebox{\linewidth}{!}{%
\begin{tabular}{ccccc}
\hline
\multicolumn{3}{c}{Components} & \multirow{2}{*}{Finetune} & \multirow{2}{*}{Linear} \\[-0.07cm]
DC      & FS        & MD        &         &       \\\hline
-           & -         & -         & 80.8    & 68.7  \\
\Checkmark  &           &           & 79.7    & 69.7  \\
\Checkmark  & \Checkmark&           & 81.3    & 71.2  \\
\Checkmark  &           & \Checkmark& 82.3    & 71.0  \\
\Checkmark  & \Checkmark& \Checkmark& \textbf{83.1}    & \textbf{73.1}  \\\hline
\end{tabular}%
}
\end{table}
%---------------------------------------------------

%--------------TABLE ABLATION STUDY ON HYPERPARAMETER---------
\begin{table*}[tbp]
\normalsize
\renewcommand\arraystretch{1.05}
\centering
\definecolor{lightgray}{RGB}{231,230,230}
\caption{Ablation studies on the hyperparametrs and key details. We report fine-tuning (ft) and linear probing (lin) accuracy on UCF101 split 1 unless otherwise specified. Default settings are colored in \sethlcolor{lightgray} \hl{gray}. }
\label{Ablation}
\begin{minipage}{0.3\textwidth}
\begin{subtable}{\linewidth}
\centering
    \caption{Distance threshold $\mathcal{T}$. $\mathcal{T}=0.7$ yields better performance in general.}
    \label{DistThre}
    \begin{tabular}{ccc}
    threshold & ft   & lin  \\ \specialrule{0.7pt}{0pt}{0pt}%\hline
    0.35      & 82.6 & 71.9 \\
    0.7       & \cellcolor[HTML]{E7E6E6}83.1 & \cellcolor[HTML]{E7E6E6}\textbf{73.1} \\
    1.4       & \textbf{83.4} & 71.0 \\
    2.8       & 80.6 & 69.8 
    \end{tabular}%
    
\end{subtable}%
\end{minipage}
\hfill
\begin{minipage}{0.3\textwidth}
\begin{subtable}{\linewidth}
\centering
    \caption{Motion decoder depth. 2 blocks of decoder is the optimal setting.}
    \label{DecoderLayer}
    \begin{tabular}{ccc}
    blocks & ft   & lin  \\ \specialrule{0.7pt}{0pt}{0pt}
    1      & 81.8 & 71.2 \\
    2      & \cellcolor[HTML]{E7E6E6}\textbf{83.1} & \cellcolor[HTML]{E7E6E6}\textbf{73.1} \\
    3      & 82.7 & 71.9 \\
           &      &         
    \end{tabular}%
    
\end{subtable}%
\end{minipage}
\hfill
\begin{minipage}{0.3\textwidth}
\begin{subtable}{\linewidth}
\centering
    \caption{Decoder width and number of heads. Excess attention heads introduce noise.}
    \label{DecoderWidth}
    \begin{tabular}{cccc}
    dim          & nheads & ft                          & lin                          \\ \specialrule{0.7pt}{0pt}{0pt}
    256          & 4     & 82.7                         & \textbf{73.3}                         \\
                 & 8     & 81.3                         & 69.4                         \\
    512          & 4     & \cellcolor[HTML]{E7E6E6}\textbf{83.1} & \cellcolor[HTML]{E7E6E6}73.1 \\
                 & 8     & 81.7                         & 71.6                         
    \end{tabular}%
    
\end{subtable}%
\end{minipage}
\hfill
\begin{minipage}{0.3\textwidth}
\begin{subtable}{\linewidth}
\centering
    % \caption{Foreground ratio $\beta$. Small $\beta$ is beneficial for the transferability of features}
    % \label{ForeRatio}
    % \begin{tabular}{cccc}
    % ratio            & ft   & lin  & ft\_hmdb   \\ \specialrule{0.7pt}{0pt}{0pt}
    % 0.3              & 82.2 & 71.6 & \textbf{56.0}      \\
    % 0.5              & \cellcolor[HTML]{E7E6E6}\textbf{83.1} & \cellcolor[HTML]{E7E6E6}\textbf{73.1} & \cellcolor[HTML]{E7E6E6}54.2    \\
    % 0.7              & 81.9 & 71.5 & 54.3     \\
    %                  &      &      &           
    % \end{tabular}%
    \caption{Foreground ratio $\beta$. Small $\beta$ is beneficial for the transferability of features}
    \label{ForeRatio}
    \begin{tabular}{ccccc}
    ratio            & ft   & lin  & ft$_{h}$ & lin$_{h}$   \\ \specialrule{0.7pt}{0pt}{0pt}
    0.3              & 82.2 & 71.6 & \textbf{56.0} & \textbf{41.2}    \\
    0.5              & \cellcolor[HTML]{E7E6E6}\textbf{83.1} & \cellcolor[HTML]{E7E6E6}\textbf{73.1} & \cellcolor[HTML]{E7E6E6}54.2 & \cellcolor[HTML]{E7E6E6}39.2  \\
    0.7              & 81.9 & 71.5 & 54.3 & 39.0   \\
                     &      &      &      &        
    \end{tabular}%
    
\end{subtable}%
\end{minipage}
\hfill
\begin{minipage}{0.3\textwidth}
\begin{subtable}{\linewidth}
\centering
    \caption{Foreground mask source. Using both modalities provides a more precise mask.}
    \label{MaskSource}
    \begin{tabular}{ccc}
    source          & ft   & lin  \\ \specialrule{0.7pt}{0pt}{0pt}
    RGB             & 81.6 & 72.1 \\
    frame difference& 82.7 & 71.1 \\
    both            & \cellcolor[HTML]{E7E6E6}\textbf{83.1} & \cellcolor[HTML]{E7E6E6}\textbf{73.1} \\
                    &      &     
    \end{tabular}%
    
\end{subtable}%
\end{minipage}
\hfill
\begin{minipage}{0.3\textwidth}
\begin{subtable}{\linewidth}
\centering
    \caption{Foreground mask position. Filtering out the noise on both sides is important.}
    \label{MaskPosition}
    \begin{tabular}{ccc}
    position        & ft   & lin  \\ \specialrule{0.7pt}{0pt}{0pt}
    no mask         & 82.3 & 71.0 \\
    prediction side & 81.8 & 71.7 \\
    target side     & 81.9 & 72.2 \\
    both            & \cellcolor[HTML]{E7E6E6} \textbf{83.1} & \cellcolor[HTML]{E7E6E6} \textbf{73.1}
    \end{tabular}%
\end{subtable}%
\end{minipage}
\end{table*}
%----------------------------------------------------------------------

\subsection{Ablation Study}

In this subsection, we perform in-depth ablation studies of FIMA. We pre-trained on split 1 of UCF101 with I3D for 200 epochs. Unless otherwise specified, we report the linear probing and fine-tuning Top-1 classification accuracy on UCF101 split 1.

\noindent
{\bf Overall Framework.} We analyze how each loss function contributes to the overall learning objective. We show the results of linear probing and fine-tuning accuracies on UCF101 and HMDB51 in Table~\ref{AblationLoss}. The vanilla instance-level video contrastive loss $\mathcal{L}_{\mathrm{VV}}$ serves as the baseline. We can observe that our pixel-level motion contrastive loss $\mathcal{L}_{\mathrm{Pix}}$ improves the baseline by a large margin and significantly outperforms the instance-level loss $\mathcal{L}_{\mathrm{VD}}$. This observation verifies our idea of eliminating the weak alignment. The frame-level local motion loss $\mathcal{L}_{\mathrm{Fra}}$ also leads to notable performance gains. It is complementary to $\mathcal{L}_{\mathrm{Pix}}$ since it improves temporal diversity by aligning motion features at the frame level.

\noindent
{\bf Components of $\mathcal{L}_{\mathrm{Pix}}$.} To eliminate the weak alignment of motion features, we propose a dense contrastive learning framework, foreground sampling strategy, and motion decoder in $\mathcal{L}_{\mathrm{Pix}}$. We ablate the effectiveness of these components in Table~\ref{AblationComponents}. When only adopting the dense contrastive framework, the performances are compromised on the classification task~\cite{DetCo}. The foreground sampling strategy and motion decoder can boost the performances independently or cooperatively by eliminating the spatial and temporal weak alignment. This also proves the existence of two kinds of weak alignment and the effectiveness of the proposed designs.

\noindent
{\bf Distance Threshold $\mathcal{T}$.} Table~\ref{DistThre} ablates the distance threshold $\mathcal{T}$ in dense contrastive learning framework. This parameter describes the range of motion features as the contrast target of a pixel. $\mathcal{T}=0.7$ yields better performance in general. The result is in accordance with the one in~\cite{PixPro}.

\noindent
{\bf Motion Decoder Design.} We first conduct experiments with different decoder depths in Table~\ref{DecoderLayer}. A 2-layer shallow motion decoder achieves the best results. More layers lead to a decrease in the results. We reason that more decoder layers may lead to overfitting of model training on the small-scale UCF101 dataset.

In Table~\ref{DecoderWidth} we ablate the decoder width and the number of heads. We observe that 8 attention heads decrease the performances. We argue that excess attention heads may sample information from some noisy latent subspaces. For the decoder width, considering that the motion decoder is also responsible for local motion reconstruction task, we use 512 dimensions by default. We provide more ablation studies in the supplementary material.

\noindent
{\bf Foreground Sampling Strategy.} We study the influence of the foreground ratio $\beta$ in Table~\ref{ForeRatio}. We additionally report the results on HMDB51 in this study, noted as ft$_h$ and lin$_h$. An intriguing observation is that $\beta=0.3$ obtains better results on HMDB51 and $\beta=0.5$ performs best on UCF101. We argue that aligning motion features with a small foreground ratio introduces the most relevant and noise-less motion information, which is critical for the transferability of the learned representations. On the other hand, a relatively large foreground ratio can provide more cues for instance discrimination but inevitably introduces more noise.

Table~\ref{MaskSource} studies the source of the foreground sampling mask. Using the combination of RGB and frame difference achieves the best results, as it locates the foreground region more precisely.

Table~\ref{MaskPosition} studies the position of the foreground sampling mask. Applying a foreground mask on the prediction or target side means filtering out the background feature pixels in the corresponding feature map. We find background noise on either the prediction or the target side could damage the learned representations. Thus sampling foreground features on both sides is important.

\begin{table*}[htbp]
\tiny
\renewcommand\arraystretch{1.05}
\centering
\caption{Action recognition performance on UCF101 and HMDB51 under linear probing and fine-tuning settings. $\dag$ denotes our reproduced results that strictly follow the settings in the original paper.}
\label{ARMain}
\resizebox{\textwidth}{!}{%
\begin{tabular}{lllllcccc}
\hline %\specialrule{0.3pt}{}{}
\multirow{2}{*}[-1.0pt]{Method} &
  \multirow{2}{*}[-1.0pt]{Backbone} &
  \multirow{2}{*}[-1.0pt]{Pretrain Dataset} &
  \multirow{2}{*}[-1.0pt]{Feames} &
  \multirow{2}{*}[-1.0pt]{Res} &
  \multicolumn{2}{c}{Linear} &
  \multicolumn{2}{c}{Finetune} \\[-0.06cm] \cmidrule(lr){6-7}\cmidrule(lr){8-9} \\[-0.3cm]
           &         &        &    &     & UCF101 & HMDB51 & UCF101 & HMDB51 \\ \hline
VCP~\cite{VCP}        & R3D-18  & UCF101 & 16 & 112 & -      & -      & 66.3   & 32.2   \\
PRP~\cite{PRP}        & R(2+1)D & UCF101 & 16 & 112 & -      & -      & 72.1   & 35.0   \\
% MLFO~\cite{MultiLevel}       & R3D-18  & UCF101 & 16 & 112 & -      & -      & 76.2   & 41.1   \\
% TCLR~\cite{TCLR}       & R(2+1)D & UCF101 & 16 & 112 & 69.9   & -      & 82.8   & 53.6   \\
DCLR~\cite{Dual}       & R(2+1)D & UCF101 & 16 & 112 & 67.1   & 40.1   & 82.3   & 50.1   \\
SDC~\cite{Static}        & R(2+1)D & UCF101 & 16 & 112 & 67.4   & 40.7   & 82.1   & 49.7   \\
\textbf{FIMA(ours)} & R(2+1)D & UCF101 & 16 & 112 & \textbf{71.2}   & \textbf{41.1}   & \textbf{84.1}   & \textbf{56.0}   \\
% \hdashline[2pt/4pt]
% \specialrule{0pt}{0pt}{0.5pt}
\specialrule{0.2pt}{0pt}{0.5pt}
BE~\cite{BE}         & I3D     & UCF101 & 16 & 224 & -      & -      & 82.4   & 52.9   \\
FAME~\cite{FAME}       & I3D     & UCF101 & 16 & 224 & \;~67.2$\dag$   & \;~36.9$\dag$   & 81.2   & 52.6   \\ 
\textbf{FIMA(ours)} & I3D     & UCF101 & 16 & 224 & \textbf{75.3}   & \textbf{42.8}   & \textbf{84.2}   & \textbf{57.8}   
\\ \specialrule{0.25pt}{0pt}{1.1pt} \specialrule{0.25pt}{0pt}{0.5pt}
CCL~\cite{CCL}        & R3D-18  & Kinetics-400 & 16 & 112 & 52.1   & 27.8   & 69.4   & 37.8   \\
% Pace~\cite{Pace}       & R(2+1)D & Kinetics-400 & 16 & 112 & -      & -      & 77.1   & 36.6   \\
MemDPC~\cite{MemoryAug}     & R3D-34  & Kinetics-400 & 40 & 224 & 54.1   & 30.5   & 78.1   & 41.2   \\
% TempTrans~\cite{TempTrans}  & R3D-18  & Kinetics-400 & 16 & 112 & -      & -      & 79.3   & 49.8   \\
RSPNet~\cite{RSPNet}     & R(2+1)D & Kinetics-400 & 16 & 112 & 61.8   & 42.8   & 81.1   & 44.6   \\
LSFD~\cite{LongShort}       & R3D-18  & Kinetics-400 & 32 & 128 & -      & -      & 77.2   & 53.7   \\
MLFO~\cite{MultiLevel}       & R3D-18  & Kinetics-400 & 16 & 112 & 63.2   & 33.4   & 79.1   & 47.6   \\
VideoMoCo~\cite{VideoMoCo}  & R(2+1)D & Kinetics-400 & 32 & 112 & -      & -      & 78.7   & 49.2   \\
TCLR~\cite{TCLR}       & R(2+1)D & Kinetics-400 & 16 & 112 & -      & -      & 84.3   & 54.2   \\
DCLR~\cite{Dual}       & R(2+1)D & Kinetics-400 & 16 & 112 & 72.3   & \textbf{46.4}   & 83.3   & 52.7   \\
FAME~\cite{FAME}       & R(2+1)D & Kinetics-400 & 16 & 112 & 72.2   & 42.2   & 84.8   & 53.5   \\
SDC~\cite{Static}        & R(2+1)D & Kinetics-400 & 16 & 112 & 72.1   & 45.9   & 86.1   & 54.8   \\
\textbf{FIMA(ours)} & R(2+1)D & Kinetics-400 & 16 & 112 & \textbf{73.1}   & 45.5   & \textbf{86.7}   & \textbf{59.4}   \\
% \hdashline[2pt/4pt]
% \specialrule{0pt}{0pt}{0.5pt}
\specialrule{0.2pt}{0pt}{0.5pt}
DSM~\cite{DSM}        & I3D     & Kinetics-400 & 16 & 224 & -      & -      & 74.8   & 52.5   \\
BE~\cite{BE}        & I3D     & Kinetics-400 & 16 & 224 & -      & -      & 86.8   & 55.4   \\
FAME~\cite{FAME}       & I3D     & Kinetics-400 & 16 & 224 & \;~75.3$\dag$ & \;~46.7$\dag$  & \textbf{88.6}   & 61.1   \\
\textbf{FIMA(ours)} & I3D     & Kinetics-400 & 16 & 224 & \textbf{76.4}   & \textbf{47.3}   & 88.5   & \textbf{62.1}   \\ \hline
\end{tabular}%
}
\end{table*}

%----------TABLE ON Diving48-----------
\begin{table}[tbp]
\normalsize
\renewcommand\arraystretch{1.0}
\setlength{\tabcolsep}{5.6mm}
\centering
\caption{Action recognition performance on Diving-48. All models are pre-trained on Kinetics-400.}
\label{ARDiving}
\resizebox{\linewidth}{!}{%
\begin{tabular}{cccc} \hline
Method        & Backbone    & Res. & Finetune  \\ \specialrule{0.7pt}{0pt}{0pt}
TCLR~\cite{TCLR}      &  R3D-18  & 112 & 22.9 \\
BE~\cite{BE}        &  I3D     & 224 & 62.4  \\
FAME~\cite{FAME}      &  I3D     & 224 & 72.9  \\
\textbf{FIMA(ours)}&  R(2+1)D & 112 & \textbf{74.7}  \\ \hline
\end{tabular}%
}
\end{table}
%-------------------------------------

\subsection{Evaluation on Downstream Tasks}

\noindent
{\bf Action Recognition on UCF101 and HMDB51.} We compare our method with the state-of-the-art self-supervised learning works on action recognition in Table~\ref{ARMain}. We report linear probing and fine-tuning Top-1 accuracy and list the detailed settings such as backbone architecture, number of frames, and resolution. For a fair comparison, we do not report methods using a deeper backbone or other modalities such as optical flow, audio, and text.

In linear probing settings, our method achieves the best results on both datasets. As the major counterparts with R(2+1)D backbone, DCLR~\cite{Dual} and SDC~\cite{Static} also utilize frame difference as the source of motion information. FIMA outperforms DCLR and SDC in general, which implies we align the features of frame difference more precisely. With the I3D backbone pre-trained on Kinetics-400 for 100 epochs, our method consistently surpasses FAME~\cite{FAME} on both UCF101 and HMDB51, which is pre-trained for 200 epochs on Kinetics-400. It suggests that explicitly incorporating motion information is more effective than in an implicit manner.

In fine-tuning settings, our method with R(2+1)D obtains state-of-the-art results on both datasets. Remarkably, our R(2+1)D model pre-trained on UCF101 gets 56.0\% classification accuracy on HMDB51, which outperforms all existing methods pre-trained on Kinetics-400. It demonstrates the data efficiency of our method and the high transferability of the learned representations. For the I3D network, FIMA pre-trained on UCF101 outperforms FAME by 3.0\% and 5.2\% on two datasets. When conducting pre-training on Kinetics-400, FIMA achieves competitive results with FAME with half training epochs (100 epochs vs. 200 epochs). 
Additionally, we report fine-tuning results on the less biased Diving-48 dataset~\cite{Diving48} in Table~\ref{ARDiving}. Our R(2+1)D pre-trained model with $112 \times 112$ resolution outperforms previous methods with a larger backbone. It demonstrates that our method introduces truly aligned motion features and effectively suppresses background bias.

%---------------------------------------------------
\begin{table}[htbp]
\Huge
\renewcommand\arraystretch{1.2}
\centering
\caption{Recall-at-topK(\%). Video retrieval performance under different K values on UCF101 and HMDB51. $\dag$ denotes our reproduced results.}
\label{VRMain}
\resizebox{\linewidth}{!}{%
\begin{tabular}{llcccccccc}
\hline
\multirow{2}{*}{Method} & \multirow{2}{*}{Backbone} & \multicolumn{4}{c}{UCF101} & \multicolumn{4}{c}{HMDB51} \\ \cline{3-10} 
           &         & R@1  & R@5  & R@10 & R@20 & R@1  & R@5  & R@10 & R@20 \\ \hline
% VCP~\cite{VCP}        & R(2+1)D & 19.9 & 33.7 & 42.0 & 50.5 & 6.7  & 21.3 & 32.7 & 49.2 \\
Pace~\cite{Pace}       & R(2+1)D & 25.6 & 42.7 & 51.3 & 61.3 & 12.9 & 31.6 & 43.2 & 58.0 \\
MLFO~\cite{MultiLevel}       & R3D-18  & 39.6 & 57.6 & 69.2 & 78.0 & 18.8 & 39.2 & 51.0 & 63.7 \\
CACL~\cite{CACL}       & R(2+1)D & 41.5 & 59.7 & 68.4 & 77.6 & 16.4 & 38.0 & 49.6 & 63.4 \\
DCLR~\cite{Dual}       & R(2+1)D & \textbf{54.8} & 68.3 & 75.9 & 82.8 & 24.1 & 44.5 & 53.7 & 64.5 \\
\textbf{FIMA(ours)} & R(2+1)D & 52.2 & \textbf{68.8} & \textbf{77.0} & \textbf{84.1} & \textbf{24.2} & \textbf{46.4} & \textbf{59.4} & \textbf{72.2} \\
% \hdashline[8pt/8pt]
% \specialrule{0pt}{0pt}{0.5pt}
\specialrule{0.25pt}{0pt}{0.5pt}
DSM~\cite{DSM}        & I3D     & 17.4 & 35.2 & 45.3 & 57.8 & 7.6  & 23.3 & 36.5 & 52.5 \\
FAME$\dag$~\cite{FAME}   & I3D     & 52.8 & 67.9 & 75.9 & 82.3 & 20.7    & 43.3    & 56.4    & 69.7    \\
\textbf{FIMA(ours)} & I3D     & \textbf{54.0} & \textbf{69.4} & \textbf{77.1} & \textbf{84.8} & \textbf{24.5} & \textbf{48.7} & \textbf{59.5} & \textbf{72.6} \\ \hline
\end{tabular}%
}
\end{table}

\noindent
{\bf Video Retrieval.} We show the video retrieval performance on UCF101 and HMDB51 in Table~\ref{VRMain}. All models are pre-trained on UCF101 with a resolution of $112 \times 112$ for R(2+1)D and $224 \times 224$ for I3D. For R(2+1)D backbone, our method generally performs better than prior work DCLR~\cite{Dual} but slightly worse in the R@1 metric on UCF101. For the I3D backbone, FIMA achieves superior results on both datasets. The stable performance improvements with different network architectures demonstrate the effectiveness and strong generalization of our method.

%------------------------------------------------------------------------
\begin{figure}[tbp]
\begin{center}
\includegraphics[width=\linewidth]{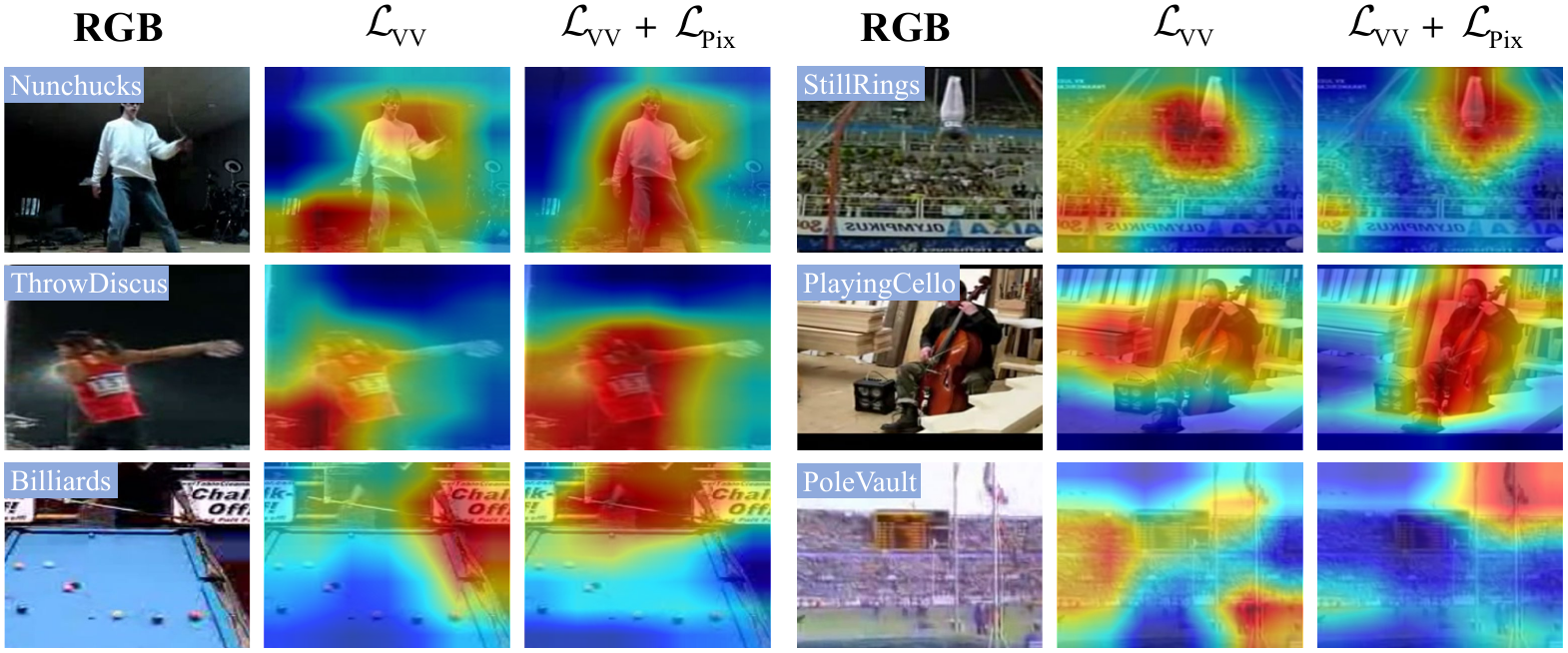}
\end{center}
% \vspace{-0.4cm}
\caption{Class-agnostic activation map visualization for MoCo baseline (middle column) and MoCo+$\mathcal{L}_{\mathrm{Pix}}$ (right column). $\mathcal{L}_{\mathrm{Pix}}$ is effective for alleviating background bias.
}
\label{CAAM_Base}
\end{figure}
%-------------------------------------------------------------------------
% \subsection{Visualization Analysis (optional)}

\subsection{Visualization of Model Attention} 
To demonstrate the effectiveness of the proposed pixel-level motion contrastive loss $\mathcal{L}_{\mathrm{Pix}}$ for alleviating the background bias, we adopt the class-agnostic activation map~\cite{CAAM} to visualize the model attention. The qualitative results of the I3D model are presented in Fig.~\ref{CAAM_Base}. We can observe that the model pre-trained with the vanilla contrastive loss $\mathcal{L}_{\mathrm{VV}}$ severely suffers from background bias and falsely attends to static cues. Since $\mathcal{L}_{\mathrm{Pix}}$ introduces fine-grained motion information, the model pre-trained with $\mathcal{L}_{\mathrm{VV}} + \mathcal{L}_{\mathrm{Pix}}$ correctly concentrates on the foreground area with significant motion.

\subsection{Why Is \texorpdfstring{$\mathcal{L}_{\mathrm{Pix}}$}{} More Effective?} 
\begin{figure}[tbp]
\begin{center}
\includegraphics[width=\linewidth]{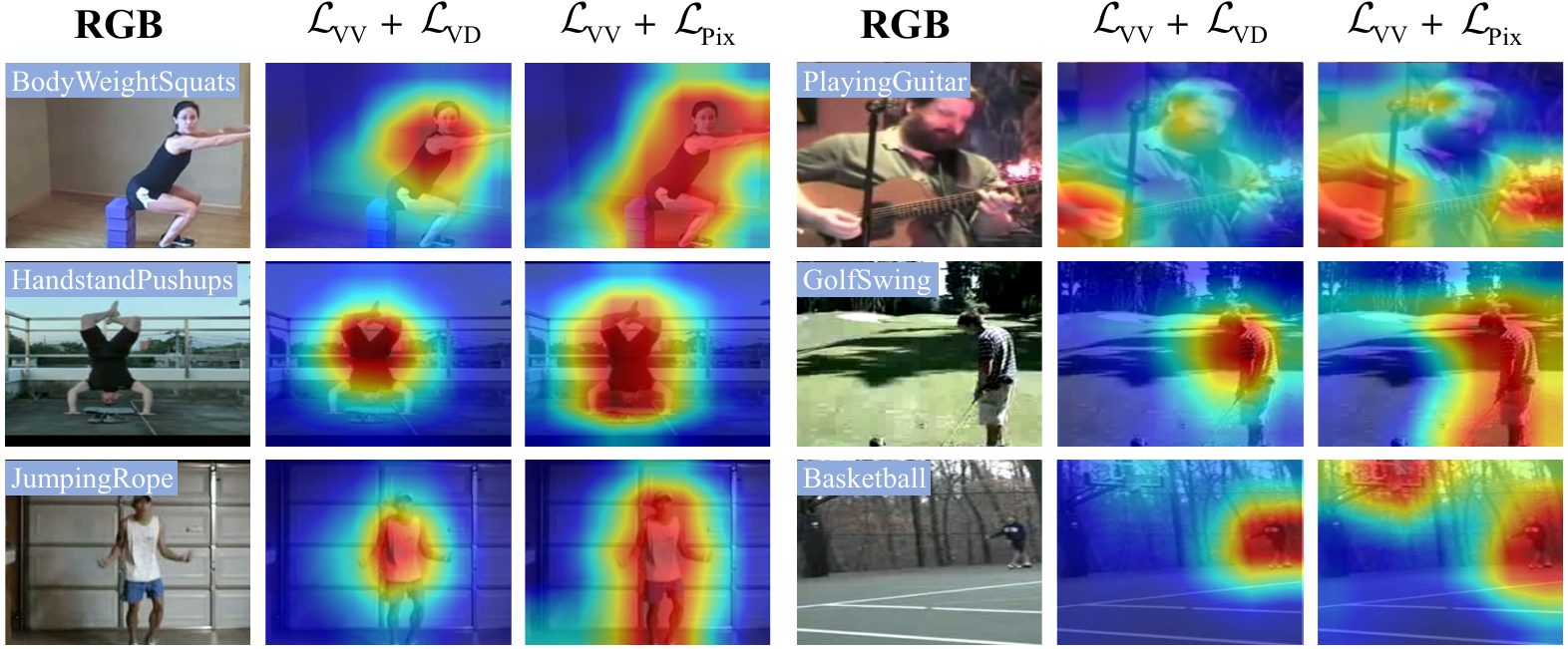}
\end{center}
% \vspace{-0.4cm}
\caption{Class-agnostic activation map visualization for MoCo+$\mathcal{L}_{\mathrm{VD}}$ (middle column) and MoCo+$\mathcal{L}_{\mathrm{Pix}}$ (right column). Pre-training with $\mathcal{L}_{\mathrm{Pix}}$ provides richer motion information.
}
\label{CAAM_Diff}
\end{figure}

To understand why is $\mathcal{L}_{\mathrm{Pix}}$ more effective than the vanilla motion contrastive loss $\mathcal{L}_{\mathrm{VD}}$, we pre-trained the I3D network with $\mathcal{L}_{\mathrm{VV}} + \mathcal{L}_{\mathrm{VD}}$ and $\mathcal{L}_{\mathrm{VV}} + \mathcal{L}_{\mathrm{Pix}}$, respectively. We visualize the class-agnostic activation maps in Fig.~\ref{CAAM_Diff}. It can be observed that the motion information introduced by $\mathcal{L}_{\mathrm{VD}}$ limited to the local area, while some significant motion cues are neglected. For example, in row-1 on the left, the ground-truth label is "BodyWeightSquats". The attention of the model trained with $\mathcal{L}_{\mathrm{VV}} + \mathcal{L}_{\mathrm{VD}}$ concentrates on the upper part of the human body. However, the movement of the legs is critical to discriminate the "BodyWeightSquats" from other motions. On the contrary, the attention of the model trained with $\mathcal{L}_{\mathrm{VV}} + \mathcal{L}_{\mathrm{Pix}}$ covers the whole foreground area, thus providing more comprehensive and holistic motion details for downstream tasks.

% \subsection{Visualization of Affinity Matrix} 
\subsection{Are Motion Features Well Aligned?} 
\begin{figure}[tbp]
\begin{center}
\includegraphics[width=\linewidth]{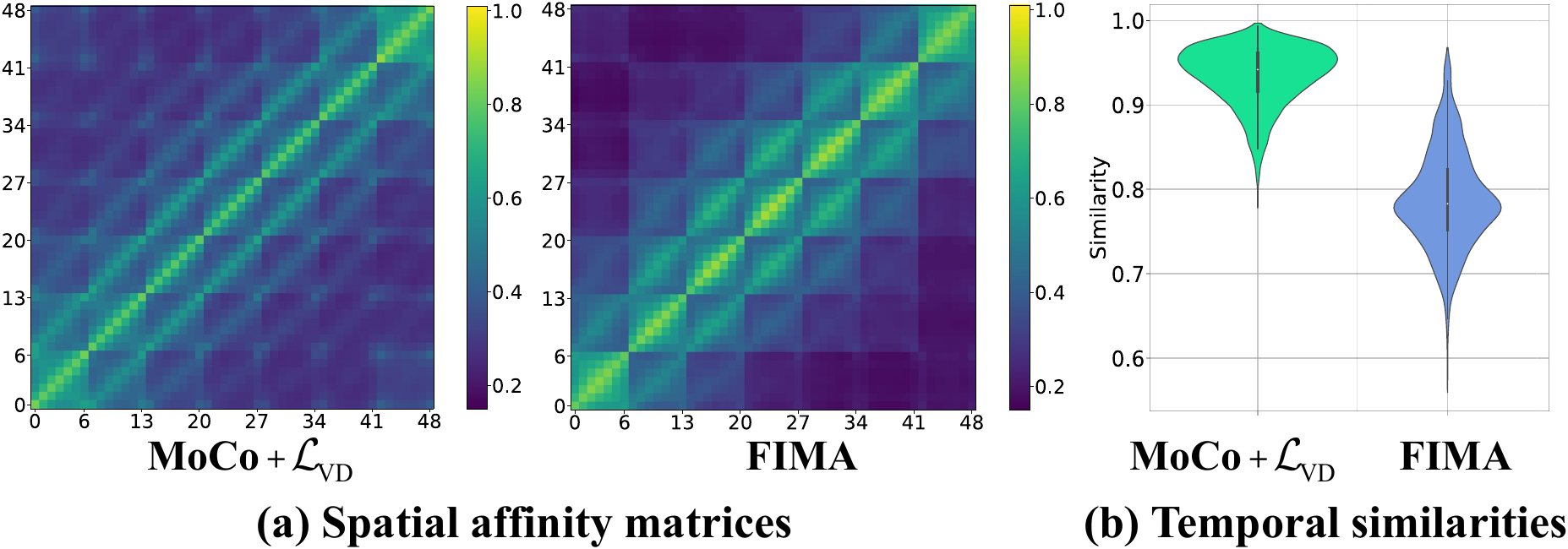}
\end{center}
% \vspace{-0.4cm}
\caption{(a) Spatial affinity matrices and (b) Temporal similarity statistics between RGB features and motion features with MoCo+$\mathcal{L}_{\mathrm{VD}}$ pre-training and FIMA pre-training.
}
\label{AM_VP}
\end{figure}

To verify whether the motion features are well aligned, we visualize the affinity matrices that describe the pairwise relationships between RGB and motion feature pixels. Specifically, we apply the average pooling to the extracted feature maps along the time dimension and then flat the output to the shape of $49 \times 1024$. Each pixel feature is normalized and the cosine similarity is used to calculate the relationship. The final matrix is averaged over 50 randomly selected video clips in the test set of UCF101. As shown in Fig.~\ref{AM_VP} (a), FIMA aligns the features of RGB and frame difference better in two aspects. First, the affinity matrix of FIMA has higher similarity around the diagonal. This indicates that every feature pixel learned by FIMA encodes more motion information around itself. Second, there are some outliers outside the diagonal in the affinity matrix of MoCo+$\mathcal{L}_{\mathrm{VD}}$, while the affinity matrix of FIMA does not. This demonstrates that FIMA can retain spatial location information and facilitate a more accurate alignment of motion features. Besides, to demonstrate the effectiveness along the temporal dimension, we randomly select 50 videos in the test set of UCF101. For each video, we uniformly sample 10 clips and extract their pooled RGB and motion features. Then calculate the similarity of each pair of RGB and motion features. We visualize the similarities as a violin plot in Fig.~\ref{AM_VP} (b). We can observe that FIMA has a smaller mean similarity with a larger deviation, which indicates that the features learned by FIMA can better capture the variation of the motion semantics.

\section{Conclusion}
In this paper, we propose a novel self-supervised learning framework to incorporate well-aligned motion information. With the fine-grained motion supervision generated by the pixel-level contrastive learning paradigm, we eliminate the spatial and temporal weak alignments by designing a motion decoder and a foreground sampling strategy. Enabling a fine-grained spatiotemporal perception enhanced by accurate motion information. We achieve state-of-the-art results on standard benchmarks and extensive ablation studies demonstrate the effectiveness of our method.

\section{Acknowledgments}
This paper is supported by the National Natural Science Foundation of China under Grants (62233013, 62073245, 62173248). Suzhou Key Industry Technological Innovation-Core Technology R\&D Program (SGC2021035).

%------------------------------------------------------------------------
%------------------------------------------------------------------------
% %------------------------------------------------------------------------
% \begin{figure}[tbp]
% \begin{minipage}{0.69\linewidth}
% \centering
%     \subfloat[][Spatial affinity matrices]{\label{AffinityMatrix}\includegraphics[width=\linewidth]{Fig_AM.pdf}}
% \end{minipage}
% \hfill
% \begin{minipage}{0.3\linewidth}
% \centering
%     \subfloat[][Statistics]{\label{ViolinPlot}\includegraphics[width=\linewidth]{Fig_VP.pdf}}
% \end{minipage}
% % \vspace{-0.4cm}
% \caption{(a) Spatial affinity matrices and (b) Temporal similarity statistics between RGB features and motion features with MoCo+$\mathcal{L}_{\mathrm{VD}}$ pre-training and FIMA pre-training.
% }
% \label{affinitymap}
% \end{figure}
% %------------------------------------------------------------------------
%------------------------------------------------------------------------
\bibliographystyle{ACM-Reference-Format}
\balance
\bibliography{References}

\appendix
\clearpage
\section{More Implementation Details}

\subsection{Technical Details.}

For the details of MoCo-v2, we closely follow the settings in~\cite{MoCov2}. We use the symmetric loss~\cite{BYOL} for optimization. The temperature parameter $\tau$ is 0.1 for all losses, and the momentum parameter used to update the key encoder is 0.999. For $\mathcal{L}_{\mathrm{VV}}$ and $\mathcal{L}_{\mathrm{Fra}}$, the size of the negative queue is 2048 for UCF101, and 65536 for Kinetics-400. The negative set $\phi_n^k$ in $\mathcal{L}_{\mathrm{Pix}}$ is implemented as a queue with a size of 784 for UCF101 and 31360 for Kinetics-400. In particular, the negative pixels are from the dense feature maps in the same mini-batch when the pre-training is conducted on UCF101. (Note that in our case, the extracted motion feature map $\Tilde{M}^k \in \mathbb{R}^{2 \times 7 \times 7 \times C}$ and we set the size of mini-batch to 8).

\subsection{Motion Decoder Details.}

\begin{figure}[htbp]
\begin{center}
\includegraphics[width=0.35\linewidth]{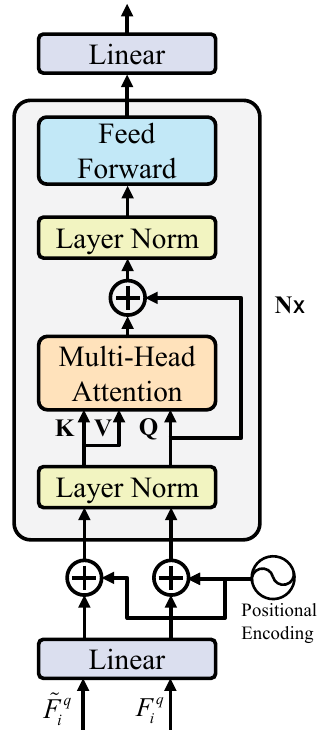}
\end{center}
% \setlength{\belowcaptionskip}{-0.3cm}
% \vspace{-0.4cm}
   \caption{The architecture of the motion decoder. }
\label{MDArch}
\end{figure}

Fig.~\ref{MDArch} shows the architecture of the motion decoder. The first and the last linear layers are used to downsize and upscale the feature dimension. We apply a dropout of 0.1 in multi-head attention and feed-forward modules. We use simple 1D absolute positional encodings since it is enough to represent the spatial relations.

\subsection{Self-supervised Pre-training Details.} 

All experiments are conducted on 3-8 NVIDIA RTX 2080 Ti GPUs, and RTX 3090 GPUs. We set mini-batch size of each GPU to 8. We use SGD optimizer with a momentum of 0.9 and a weight decay of $10^{-4}$. We randomly sample two 16-frame clips with a temporal stride of 2 at different timestamps and compute the frame difference, thus covering $16 \times 2$ frames in total. For the frame-level motion reconstruction task, the 32 frames clip is divided into 2 sub-clips, each with 16 frames and a temporal stride of 1 and then compute the frame difference.   

Following~\cite{ALargeScale,FAME}, we use geometric data augmentation with RandomResizedCrop (scale $\in [0.2, 1.0]$) and RandomHorizontalFlip (probability=0.5). For photometric augmentation, we adopt RandomGrayscale (probability=0.2), ColorJitter (probability=0.2), and RandomGaussianBlur (probability=0.2, kernel\_size= 23 with standard deviation $\in [0.1, 2.0]$).

\subsection{Details of Action Recognition.} 

In fine-tuning settings, we train for 150 epochs using the SGD optimizer with a momentum of 0.9 and a weight decay of $10^{-4}$. We set the initial learning rate to 0.0033 with a batch size of 16 and we decay the learning rate at epoch 60 and epoch 120 by a factor of 10. We use a dropout of 0.7 for UCF101 and 0.5 for HMDB51. Data augmentation is the same as in pre-training stage, except that RandomGaussianBlur is not applied.

In linear probing settings, we train for 100 epochs with an initial learning rate of 1.5 and a batch size of 16. The learning rate is decayed at epoch 60 and epoch 80 by a factor of 10. No dropout and weight decay is applied.

\section{Additional Ablation Studies}

We provide additional ablation studies in this section. The settings are the same as the main paper ablation studies.

\subsection{Decoder Width on All Losses.} 
In the main paper, we conduct ablations on the decoder width using $\mathcal{L}_{\mathrm{VV}}$ and $\mathcal{L}_{\mathrm{Pix}}$. We provide an 
 additional ablation in Table~\ref{AblationAddiDW} using all losses: $\mathcal{L}_{\mathrm{VV}}$, $\mathcal{L}_{\mathrm{Pix}}$, and $\mathcal{L}_{\mathrm{Fra}}$.

When combined with the local motion reconstruction task, the decoder width of 256 dimensions is not capable enough to optimize both $\mathcal{L}_{\mathrm{Pix}}$ and $\mathcal{L}_{\mathrm{Fra}}$, leading to a decrease in performance. Thus, we use 512 dimensions by default.

% -------------TABLE ABLATION STUDY ON Decoder Width----------------
\begin{table}[htbp]
\normalsize
\renewcommand\arraystretch{1.1}
\setlength{\tabcolsep}{10mm}
\centering
\caption{Ablation study on the decoder width using $\mathcal{L}_{\mathrm{VV}}$, $\mathcal{L}_{\mathrm{Pix}}$, and $\mathcal{L}_{\mathrm{Fra}}$.}
\label{AblationAddiDW}
\resizebox{\linewidth}{!}{%
\begin{tabular}{ccc}
dim        & ft    & lin  \\ \specialrule{0.7pt}{0pt}{0pt}
256        & \cellcolor[HTML]{FFFFFF} 82.5  & \cellcolor[HTML]{FFFFFF} 72.3 \\
512        & \cellcolor[HTML]{E7E6E6} \textbf{84.2}  & \cellcolor[HTML]{E7E6E6} \textbf{75.3}  \\
\end{tabular}%
}
\end{table}

% -------------TABLE ABLATION STUDY ON SAHRED DECODER----------------
\begin{table}[bp]
\normalsize
\renewcommand\arraystretch{1.1}
\setlength{\tabcolsep}{8mm}
\centering
\caption{Ablation study on the shared or different motion decoder.}
\label{AblationSharedDecoder}
\resizebox{\linewidth}{!}{%
\begin{tabular}{ccc}
decoder   & parameters & lin  \\ \specialrule{0.7pt}{0pt}{0pt}
shared    &  14.4M     & \cellcolor[HTML]{E7E6E6} \textbf{75.3} \\
different &  16.5M     & \cellcolor[HTML]{FFFFFF} 74.0  \\
\end{tabular}%
}
\end{table}
% --------------------------------------------------------

\subsection{Effects of Using Shared Motion Decoders.} 
In our framework, we use the shared motion decoder to reduce the computational cost. In this way, two tasks can be accomplished in a single forward process, while using different decoders requires two forward processes and adds additional training parameters. Further, using the same decoder makes the optimization of two tasks more difficult, which potentially requires a higher quality of the features extracted by the encoder. To verify this, we perform an study as shown Table~\ref{AblationSharedDecoder}. Using a shared decoder achieves better results with fewer parameters.

% -------------TABLE ABLATION STUDY ON Lfra TYPES----------------
\begin{table}[htbp]
\normalsize
\renewcommand\arraystretch{1.1}
\setlength{\tabcolsep}{15mm}
\centering
\caption{Ablation study on the various type of $\mathcal{L}_{\mathrm{Fra}}$.}
\label{AblationLfraType}
\resizebox{\linewidth}{!}{%
\begin{tabular}{cc}
loss types   &  lin  \\ \specialrule{0.7pt}{0pt}{0pt}
InfoNCE      & \cellcolor[HTML]{E7E6E6} \textbf{75.3} \\
Triplet      & \cellcolor[HTML]{FFFFFF} 73.4  \\
MSE          & \cellcolor[HTML]{FFFFFF} 71.2  \\
\end{tabular}%
}
\end{table}
% --------------------------------------------------------

\subsection{Various Types of Frame-level Motion Loss.} 
The goal of the frame-level motion feature reconstruction is to enhance the temporal diversity of the learned features. By discriminating the positive sample from the intra-video negatives and inter-video negatives, the features extracted by the encoder become more discriminative along the temporal dimension (i.e. more time-aware). Thus, the introduction of negative samples is essential. Note that there are also other types of reconstruction losses such as Triplet Loss and MSE Loss. We perform an ablation study to explore their effects in Table~\ref{AblationLfraType}. From the result, we can see that the InfoNCE loss performs best and MSE is the worst since no negative samples are introduced.

% -------------TABLE ABLATION STUDY ON Negative pool size----------------
\begin{table}[htbp]
\normalsize
\renewcommand\arraystretch{1.1}
\setlength{\tabcolsep}{7mm}
\centering
\caption{Ablation study on the number of negative samples in $\mathcal{L}_{\mathrm{Fra}}$.}
\label{AblationNumNegative}
\resizebox{\linewidth}{!}{%
\begin{tabular}{ccc}
losses   & negative samples &lin  \\ \specialrule{0.7pt}{0pt}{0pt}
$\mathcal{L}_{\mathrm{VD}}$  &  - & \cellcolor[HTML]{FFFFFF} 68.7 \\
$\mathcal{L}_{\mathrm{Pix}}$ & 98 & \cellcolor[HTML]{FFFFFF} 70.3  \\
$\mathcal{L}_{\mathrm{Pix}}$ & 784 & \cellcolor[HTML]{E7E6E6} \textbf{73.1}  \\
$\mathcal{L}_{\mathrm{Pix}}$ & 1568& \cellcolor[HTML]{FFFFFF} 72.6  \\
\end{tabular}%
}
\end{table}
% --------------------------------------------------------

\subsection{The Number of Negative Samples in Pixel-level Motion Loss.} 
We perform an ablation study about the number of negative samples in $\mathcal{L}_{\mathrm{Pix}}$ 
as shown in Table~\ref{AblationNumNegative}. We set the number of negative samples to 98 (i.e. no motion features from other videos), 784 (i.e. motion features from the same mini-batch), and 1568 (i.e. use a queue to store motion features from other videos). We find that the introduction of motion pixel features from other videos facilitates a more accurate alignment of fine-grained motion features. Specfically, the introduction of the motion features from other videos boosts the performance, and the performance saturates after a certain number.

% -------------TABLE ABLATION STUDY ON Training costs----------------
\begin{table}[htbp]
\normalsize
\renewcommand\arraystretch{1.1}
\setlength{\tabcolsep}{2mm}
\centering
\caption{Ablation study on the training costs of $\mathcal{L}_{\mathrm{Pix}}$. GPU-days is the number of GPUs used for pre-training multiplied by the training time in days.}
\label{AblationCosts}
\resizebox{\linewidth}{!}{%
\begin{tabular}{ccccccc}
losses   & DC & FS & MD & parameters & GPU-days & lin  \\ \specialrule{0.7pt}{0pt}{0pt}
$\mathcal{L}_{\mathrm{VD}}$  & - & - &  -  &  12.3M & 3.43 & \cellcolor[HTML]{FFFFFF} 68.7 \\
$\mathcal{L}_{\mathrm{Pix}}$ &\Checkmark & - & - & 12.3M & 3.53 & \cellcolor[HTML]{FFFFFF} 69.7  \\
$\mathcal{L}_{\mathrm{Pix}}$ &\Checkmark & \Checkmark & - & 12.3M & 3.54 & \cellcolor[HTML]{FFFFFF} 71.2  \\
$\mathcal{L}_{\mathrm{Pix}}$ &\Checkmark & \Checkmark & \Checkmark & 14.4M & 3.78 & \cellcolor[HTML]{E7E6E6} \textbf{73.1}  \\
\end{tabular}%
}
\end{table}
% --------------------------------------------------------

\subsection{Training Costs of Pixel-Level Motion Loss.} 
 We report the actual pre-training time of our method with respect to the baseline as shown in Table~\ref{AblationCosts}. The wall-clock time of pre-training is benchmarked in 3 2080ti GPUs with a batch size of 24. Compared to $\mathcal{L}_{\mathrm{VD}}$
, we can see that the actual computational overhead added by the dense contrastive (DC) framework is marginal (+0.1 days). This is because most of the operations of the dense contrastive framework can be handled by efficient matrix multiplication in PyTorch. The motion decoder (MD) brings a +0.24 days training time increase due to the additional parameters.

% \noindent
\section{Limitations and Future Work.}
Our method uses a non-parametric strategy and a fixed ratio $\beta$ to sample the foreground pixels. Due to the strong bias of the model at the beginning of pre-training and the varying foreground ratio across videos, it inevitably omits useful information or introduces noise. Besides, our method originates from the dense contrastive learning framework. It is non-trivial to apply our method to dense predictive tasks in the video domain.

%--------------Turn on for full appendix-------
%------------------------------------------------------------------------
\begin{figure*}[tbp]
\begin{center}
\includegraphics[width=\textwidth]{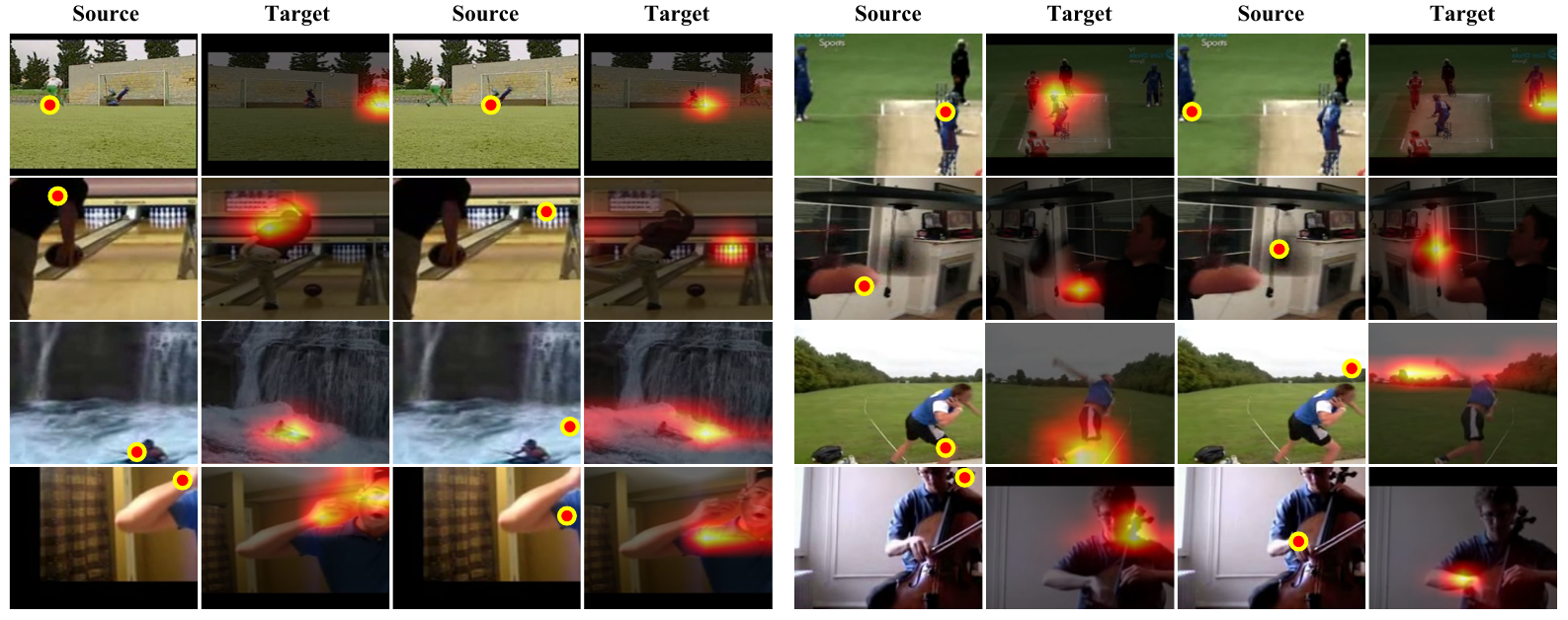}
\end{center}
% \vspace{-0.4cm}
\caption{Point attention visualization of the motion decoder. In each group, we show the original source view with two selected points denoted by the red circle (column-1 and column-3), and their corresponding attention map from the last layer of the motion decoder in the target view (column-2 and column-4). Best viewed in color and zoomed in.
}
\label{MDAttention}
\end{figure*}
%------------------------------------------------------------------------

%------------------------------------------------------------------------
\begin{figure*}[tbp]
\begin{center}
\includegraphics[width=\textwidth]{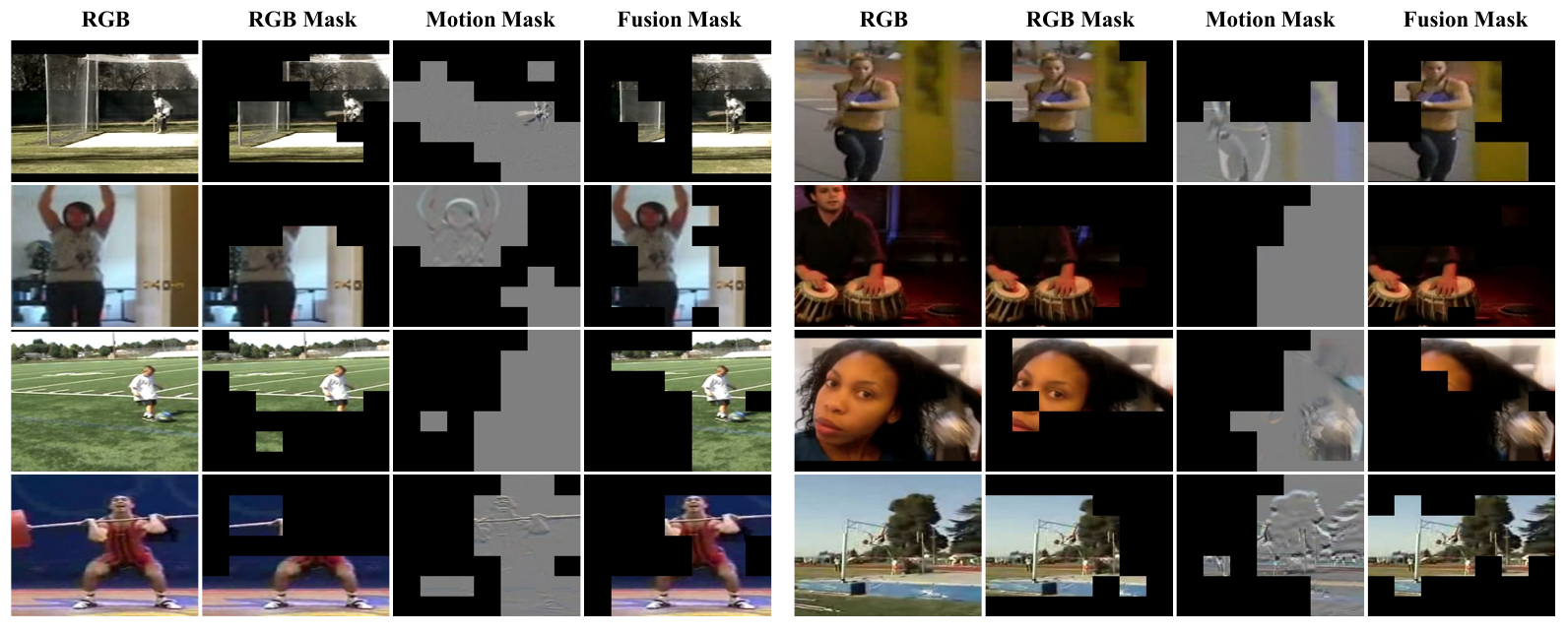}
\end{center}
% \vspace{-0.4cm}
\caption{ Visualization of the foreground sampling mask. In each group we show the RGB input (column-1), the foreground mask derived from the RGB feature map (column-2), the foreground mask derived from the frame difference feature map (column-3), and the fused foreground mask (column-4).
}
\label{foremask}
\end{figure*}
%------------------------------------------------------------------------

%------------------------------------------------------------------------
\begin{figure*}[tbp]
\begin{center}
\includegraphics[width=\textwidth]{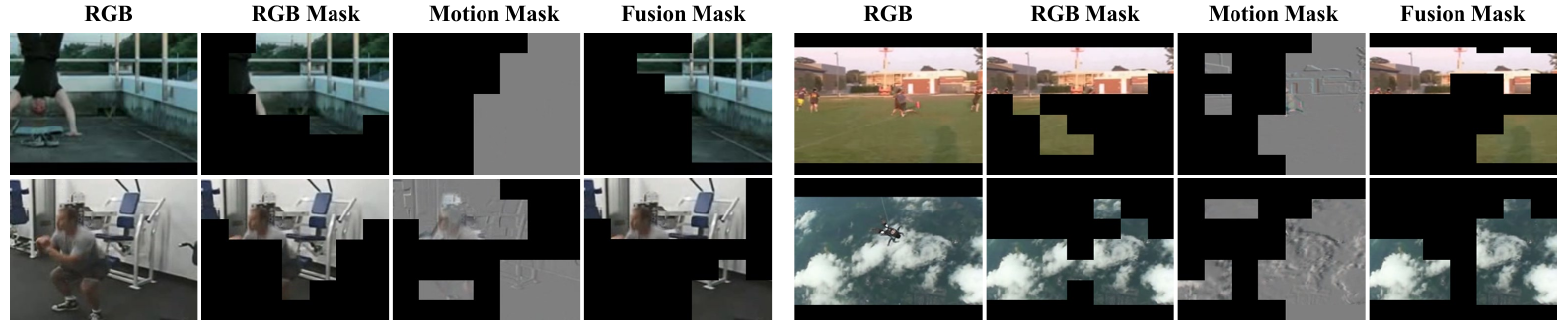}
\end{center}
% \vspace{-0.6cm}
% \setlength{\belowcaptionskip}{-0.4cm}
\caption{Failure cases of the foreground sampling mask.
}
% \vspace*{18cm}
\label{failuremask}
\end{figure*}
%------------------------------------------------------------------------

% \newpage
\section{More Visualization Analysis}

\subsection{Point Attention of Motion Decoder.}
In order to get a more intuitive understanding of how the motion decoder works, we visualize the point-attention map after pre-training in Fig.~\ref{MDAttention}. For each source view, we first randomly pick one point (denoted by the red circle). Then, the RGB feature of this point and all features of the target view are input to the motion decoder. We visualize the attention score from the last layer of the motion decoder. The score is averaged over all attention heads.

As shown in Fig.~\ref{MDAttention}, the motion decoder correctly matches the features of the person (row-1) and object (row-2), even with large variations in appearance. Further, with the input being background features, the motion decoder can also distinguish the background features in the target view (row-3). It coincides with our idea about avoiding the misalignment of background features by filtering out background features. In the last row on the left, it is interesting to see that the motion decoder can properly match features under strong occlusion. These observations suggest that the motion decoder builds an accurate and robust correspondence between the source view and the target view. These also demonstrate that the representations learned by FIMA have a clear distinction between different concepts in terms of person, object, and background.

\subsection{Foreground Sampling Mask.} 
We show the visualization of the foreground sampling mask in Fig.~\ref{foremask}. We compute class-agnostic activation maps~\cite{CAAM} of two modalities by applying the average pooling along channel and time dimensions. Then, we select patches with top-$[\beta H W]$ activation value as the foreground area. The foreground ratio $\beta$ is set to 0.5. The black patches represent the filtered-out background. We find that either RGB or frame difference feature maps are likely to be disturbed by the background region. When one of the RGB masks and the frame difference masks deviates, the fused feature map always favors the correct one. This indicates that the strategy of using both RGB and frame difference to jointly determine the foreground features is working.

\subsection{Failure Cases of Foreground Sampling.} We show some failure cases of the foreground sampling mask in Fig.~\ref{failuremask}. When neither the RGB mask nor the frame difference mask focuses on the correct foreground region, the fused feature map also fails to locate the correct foreground area.

\subsection{Visualization of Learned Representations.} 
\begin{figure*}[htbp]
\begin{center}
\includegraphics[width=0.65\textwidth]{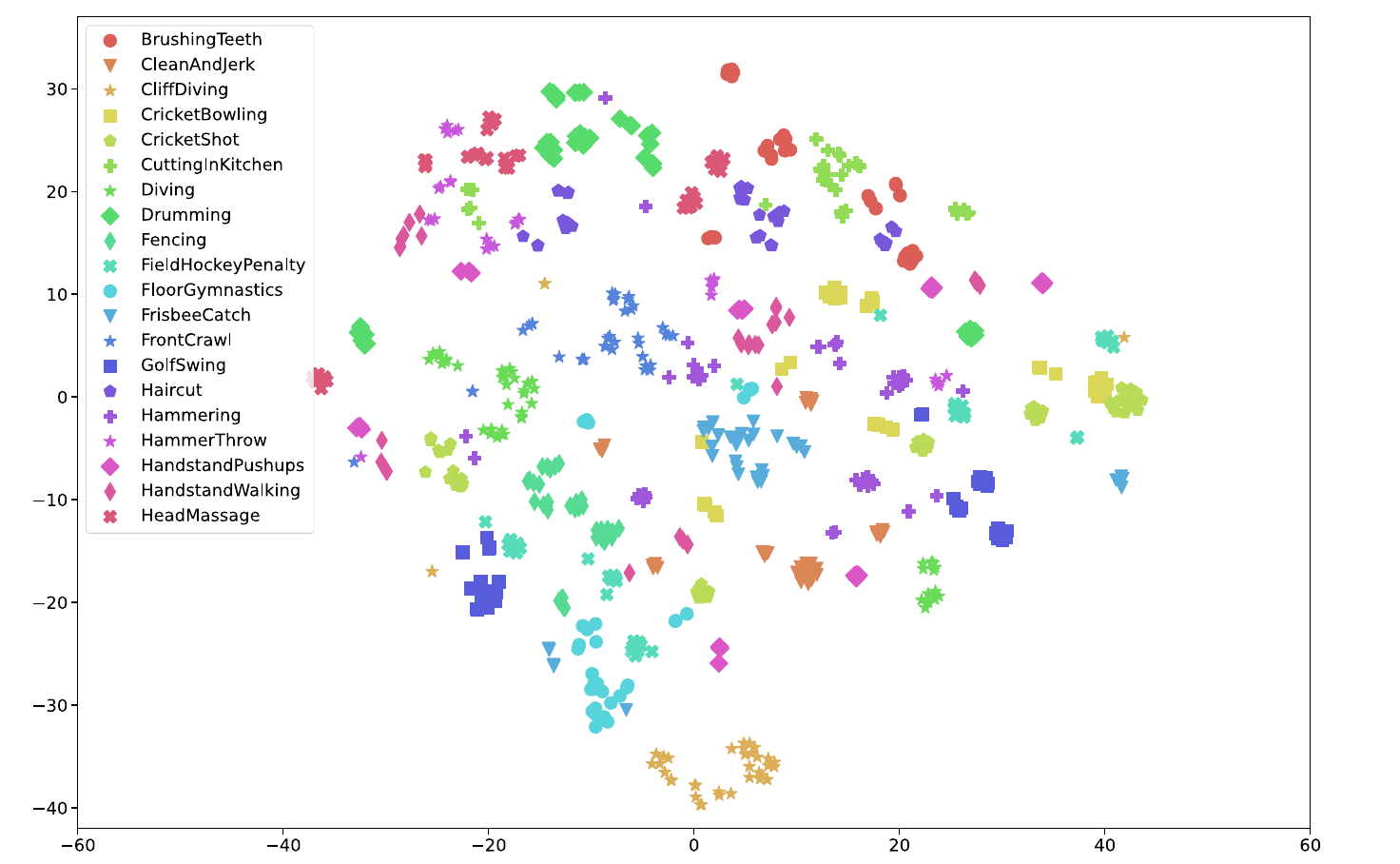}
\end{center}
% \vspace{-0.6cm}
% \setlength{\belowcaptionskip}{-0.4cm}
\caption{t-SNE visualization of FIMA representations for 20 randomly selected classes from the UCF101 test set.
}
\label{tsnefima}
\end{figure*}

\begin{figure*}[htbp]
\begin{center}
\includegraphics[width=0.65\textwidth]{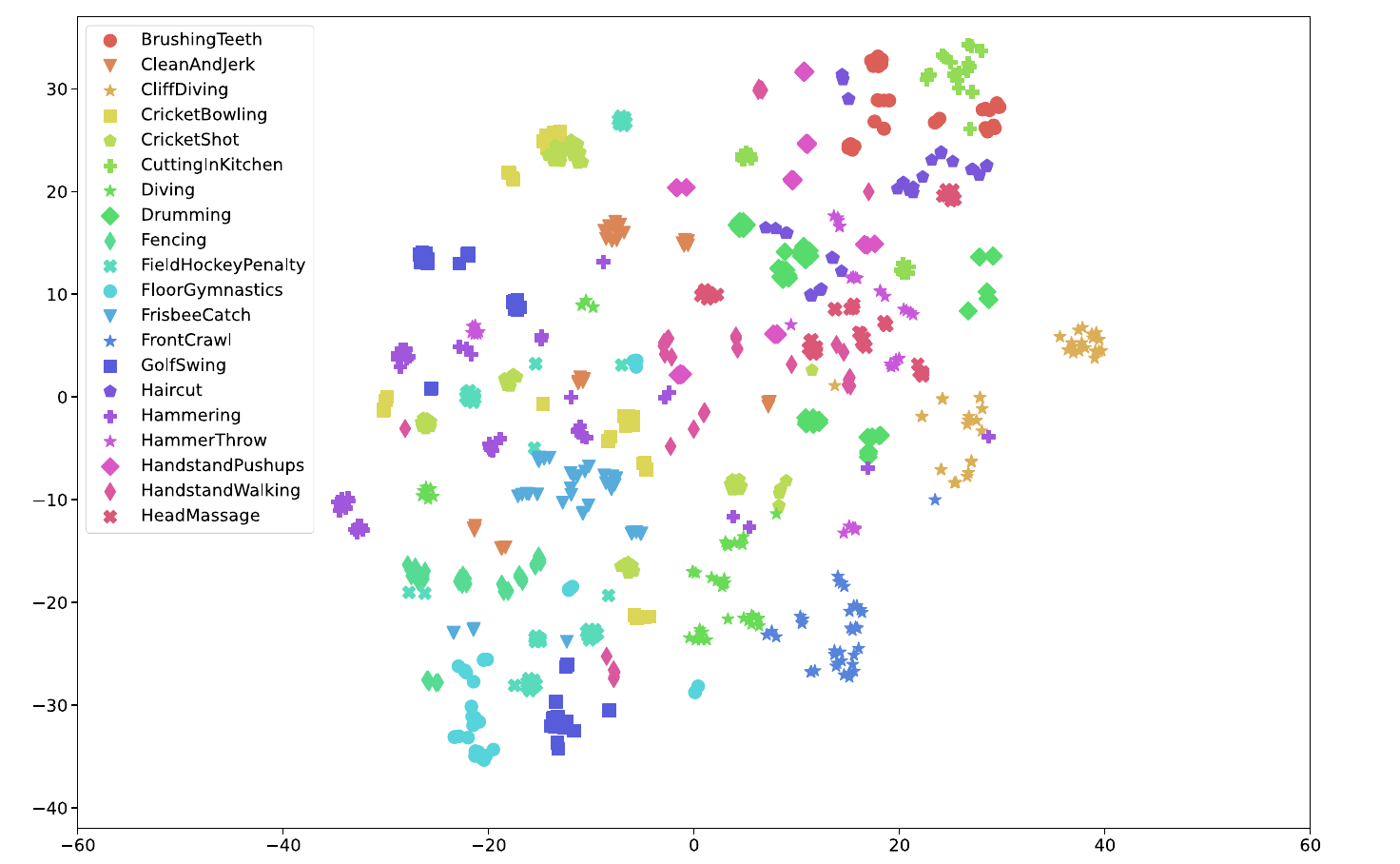}
\end{center}
% \vspace{-0.6cm}
% \setlength{\belowcaptionskip}{-0.4cm}
\caption{t-SNE visualization of MoCo representations for 20 randomly selected classes from the UCF101 test set.
}
\label{tsnemoco}
\end{figure*}

We use t-SNE~\cite{tsne} to visualize the representations learned by FIMA and MoCo baseline in Fig.~\ref{tsnefima} and Fig.~\ref{tsnemoco}. We pre-train the model on split 1 of UCF101 and visualize the representations of 20 randomly selected classes from the test set of UCF101. The perplexity for t-SNE is set to 30. From Fig.~\ref{tsnefima} and Fig.~\ref{tsnemoco}, we can observe that FIMA can facilitate the distribution of video representations in more discrete clusters.

% %------------------------------------------------------------------------
% \begin{figure}[htbp]
% \begin{minipage}{0.98\linewidth}
% \centering
%     \subfloat[][FIMA]{\label{AM_MoCo}\includegraphics[width=\linewidth]{Fig_TSNE_FIMA.pdf}}
% \end{minipage}
% % \hfill
% \begin{minipage}{0.98\linewidth}
% \centering
%     \subfloat[][MoCo]{\label{AM_FIMA}\includegraphics[width=\linewidth]{Fig_TSNE_MoCo.pdf}}
% \end{minipage}
% % \vspace{-0.4cm}
% \caption{t-SNE visualization of (a) FIMA and (b) MoCo representations for 20 randomly selected classes from the UCF101 test set.
% }
% \label{affinitymap}
% \end{figure}
% %------------------------------------------------------------------------

% %------------------------------------------------------------------------
% \begin{figure*}[htbp]
% \begin{minipage}{0.46\linewidth}
% \centering
%     \subfloat[][FIMA]{\label{AM_MoCo}\includegraphics[width=\linewidth]{Fig_TSNE_FIMA.pdf}}
% \end{minipage}
% \hfill
% \begin{minipage}{0.46\linewidth}
% \centering
%     \subfloat[][MoCo]{\label{AM_FIMA}\includegraphics[width=\linewidth]{Fig_TSNE_MoCo.pdf}}
% \end{minipage}
% % \vspace{-0.4cm}
% \caption{t-SNE visualization of (a) FIMA and (b) MoCo representations for 20 randomly selected classes from the UCF101 test set.
% }
% \label{affinitymap}
% \end{figure*}
% %------------------------------------------------------------------------

%------------------------------------------------------------------------

%------------------------------------------------------------------------

%------------------------------------------------------------------------

%------------------------------------------------------------------------

\end{document}